\documentclass[journal]{IEEEtran}
\ifCLASSINFOpdf
\else
\fi

\usepackage{graphicx}
\usepackage{amsmath}
\usepackage{amssymb}

\usepackage{caption}
\usepackage{makecell}
\usepackage{booktabs}
\usepackage{multirow}

\usepackage{multicol}
\usepackage{enumitem}

\usepackage{comment}
\usepackage{makecell}
\usepackage{booktabs}

\usepackage{url}

\usepackage{breakurl}
\usepackage[breaklinks]{hyperref}

\usepackage{algorithmic}
\usepackage{algorithm}

\usepackage{adjustbox} 

\usepackage[table,x11names,dvipsnames,xcdraw]{xcolor}
\usepackage{tabu}

\usepackage{arydshln}
\usepackage{cleveref}
\usepackage{caption}
\usepackage{subcaption}
\usepackage{booktabs}
\usepackage{tikz}
\usepackage{pgfplots}
\pgfplotsset{compat=1.9}
\usepackage{mwe}
\usepackage{graphbox}
\usepackage{soul}

\newcommand{\kr}[1]{\textcolor{red}{Kiran: #1}}

\newcommand{\re}{\textcolor{black}}
\newcommand{\rere}{\textcolor{black}}
\newcommand{\Rev}{\textcolor{black}}

\begin{document}
%
\title{Analyzing Fairness in Deepfake Detection With \\ Massively Annotated Databases}

%
%
%

\author{
Ying Xu,~\IEEEmembership{Student Member,~IEEE}, 
Philipp Terhöst,~\IEEEmembership{Member,~IEEE},\\ 
Marius Pedersen,~\IEEEmembership{Member,~IEEE},
Kiran Raja,~\IEEEmembership{Senior Member,~IEEE} 
\thanks{Ying Xu and Philipp Terhöst contributed equally to this work.}\thanks{e-mail:\{ying.xu@ntnu.no\}}
}
\maketitle

\begin{abstract}
In recent years, image and video manipulations with Deepfake have become a severe concern for security and society. Many detection models and datasets have been proposed to detect Deepfake data reliably.
However, there is an increased concern that these models and training databases might be biased and, thus, cause Deepfake detectors to fail.
In this work, we investigate \Rev{factors causing biased detection} in public Deepfake datasets by (a) \Rev{creating} large-scale demographic and non-demographic attribute annotations \Rev{with} 47 different attributes for five popular Deepfake datasets and (b) comprehensively analysing \Rev{attributes resulting in} AI-bias of three state-of-the-art Deepfake detection backbone models on these datasets.
The analysis \Rev{shows how various attributes  influence} a large variety of distinctive attributes (from over 65M labels) on the detection performance \Rev{which includes} demographic (age, gender, ethnicity) and non-demographic (hair, skin, accessories, etc.) attributes.
The results \rere{examined datasets show limited diversity} and, more importantly, show that the utilised Deepfake detection backbone models are \Rev{strongly affected by investigated attributes} making them \Rev{not fair across attributes}.
\Rev{The Deepfake detection backbone methods trained on such imbalanced/biased datasets result in incorrect detection results leading to generalisability, fairness, and security issues.}
\Rev{Our findings and annotated datasets will guide future research} to evaluate and mitigate bias in Deepfake detection techniques.
The annotated datasets \re{and the corresponding code} are publicly available\footnote{\url{https://github.com/xuyingzhongguo/DeepFakeAnnotations}}.
\end{abstract}


%
\IEEEpeerreviewmaketitle

\section{Introduction}
\label{sec:Introduction}



\IEEEPARstart{D}{eepfake} refers to a deep learning-based technique that is able to create fake videos/images by swapping the face of a person with the face of another.
\re{An example of a Deepfake would be a video of someone convincingly speaking a language they have never learned, or even impersonating a public figure so seamlessly that it's nearly impossible to distinguish from reality.
Deepfake has become a great concern for security and society \cite{DBLP:journals/inffus/TolosanaVFMO20} due to the harmful usage of such fake content, such as fake news, fake pornography, or financial fraud.}
Moreover, the availability of large-scale public face datasets and the development of strong generative artificial intelligence (AI), and especially deep learning techniques, such as Autoencoder or Generative Adversarial Networks (GAN) \cite{DBLP:conf/nips/GoodfellowPMXWOCB14,DBLP:journals/nn/Schmidhuber15} have strongly increased the realism of Deepfake.
Various open-source and mobile applications \cite{reface, faceswap} further allow to create highly realistic Deepfake videos or images without any expert knowledge and thus, make it possible for everyone to automatically manipulate images of videos with Deepfake technology.

Consequently, many works have developed detection methods capable of detecting such face manipulations \cite{DBLP:journals/mta/Zhang22a}.
Previous studies, however, pointed out some bias issues with these detection methods for different factors such as age, gender, and ethnicity \cite{hazirbas2021towards, trinh2021examination, pu2022fairness}.
The main reasons for bias in such AI models are believed to originate from unbalanced training databases \cite{hazirbas2021towards, trinh2021examination, pu2022fairness}.
Biased decisions from detection approaches \re{significantly impact} both security and society if, for example, images from a certain group of people are \re{constantly} scrutinised as Deepfake.

\subsection{\re{Societal and Technological Aspects}}
\begin{figure}[htp]
    \centering
    \includegraphics[width=1.0\linewidth]{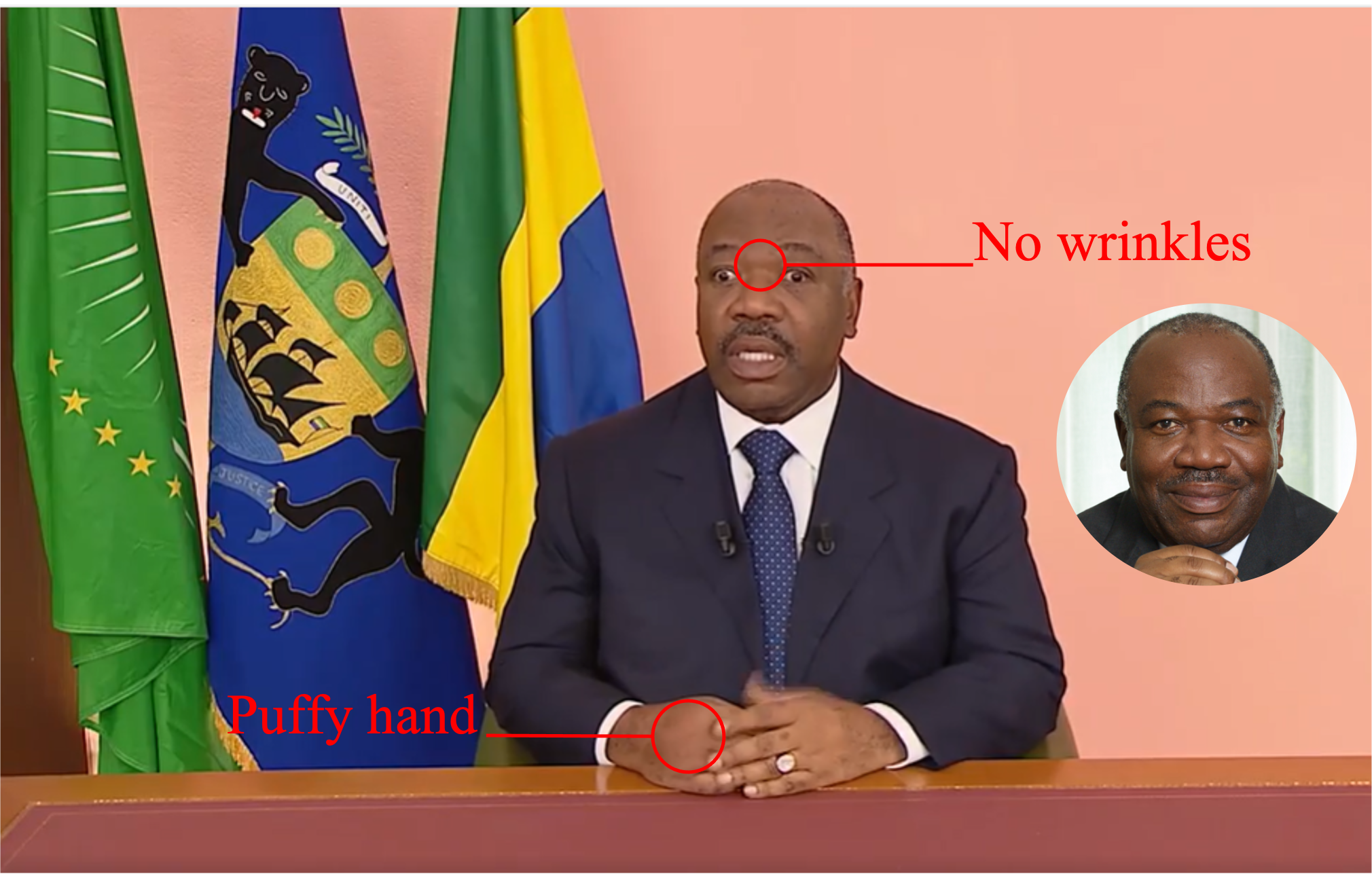}
    \caption{\Rev{Footage of Gabon's former president Ali Bongo delivering a New Year's address, with allegations suggesting it may be a deepfake~\cite{youtube}.}} 
    \label{fig:ganbo}
\end{figure}
\Rev{In 2018, the disappearance of Gabon's President Ali Bongo led to public unrest and speculation of his assassination, culminating in a government-released DeepFake New Year's broadcast that inadvertently provoked a military mutiny due to its unnatural appearance.~\cite{aligabon}. Reliable detection across different demographic groups in such situations can prevent inadvertent consequences.}
Despite the availability of DeepFake detection algorithms, there is a growing concern that these algorithms as with other machine learning algorithms,  misclassify authentic images from specific ethnic and demographic groups as DeepFakes. 

\par Such misclassification \Rev{for certain groups} can have unintended and significant societal and political consequences \Rev{\cite{raji2020saving,waelen2023struggle}}. To address this issue, it is essential to thoroughly investigate and analyse the factors related to bias in DeepFake detection. With a comprehensive understanding of these factors, the deployment of such technology can be justified \Rev{and known limitations can be disclosed}. Therefore, our primary motivation is to thoroughly investigate and analyse the factors responsible for bias in DeepFake detection.
\par \re{This paper specifically focuses on analysing these factors and does not propose or introduce any new DeepFake detection methods. Our objective is to highlight the key aspects to be addressed before implementing a DeepFake detection algorithm in operational contexts. By doing so, we aim to provide valuable insights to inform the development and deployment of more fair and accurate DeepFake detection systems.}

\subsection{Contributions}
\re{The present study highlights the necessity for annotated datasets and balanced performance metrics to assess the impact of biased datasets to determine the efficacy of detection models. In this regard, this work makes two significant contributions by analyzing factors that lead to perceived bias in Deepfake detection. }
\begin{enumerate}
    \item We provide massive and diverse annotations for five widely-used Deepfake detection datasets. Existing Deepfake detection datasets contain none or only sparse annotations restricted to demographic attributes, as shown in \Cref{tab:DatasetComparison}. This work provides over 65.3M labels using 47 different attributes for five popular Deepfake detection datasets (Celeb-DF~\cite{li2020celeb}, DeepFakeDetection (DFD)~\cite{ffdfd}, FaceForensics++ (FF++)~\cite{Rossler_2019_ICCV}, DeeperForensics-1.0 (DF-1.0)~\cite{jiang2020deeperforensics1} and Deepfake Detection Challenge Dataset (DFDC)~\cite{dolhansky2020deepfake}). 
    \item We comprehensively analyse detection bias in three state-of-the-art Deepfake detection backbone models with respect to various demographic and non-demographic attributes regarding to four current Deepfake datasets. Previous investigations restricted their analysis to a maximum of four demographic attributes on a single dataset.
    Contrarily, we analyse detection bias on a much larger scale of distinctive attributes on four widely-used Deepfake datasets\footnote{For the analysis, we do not consider DF-1.0 data as the detection methods did not produce enough errors (Details in \Cref{tab:Deeperf-effb0}\Cref{tab:Deeperf-xception}\Cref{tab:Deeperf-capnet}) on this dataset to analyse biased behaviours.}.  
\end{enumerate}

For the first contribution, five annotated datasets are created in the direction of earlier work using the MAAD-Face principle \cite{terhorst2021maad}.
By computing a reliability score from the predictions of the MAAD classifier, we consider high-confidence predictions for labelling process to ensure a high annotation correctness. 
While the annotations from previous works at most contain demographic information like age, gender, and ethnicity, the annotations in this work are highly diverse and include attributes such as hair-color and -style, skin, face geometry, mouth, noise, and various accessories.
We assert that these rich annotations will allow future works to evaluate the role of each attribute and use it to train better detection models that can mitigate bias issues.

The second contribution of our work is a detailed analysis of detection bias in Deepfake detection approaches by comparing the differential outcomes of three state-of-the-art Deepfake backbone networks (EfficientNetB0~\cite{tan2019efficientnet}, Xception~\cite{chollet2017xception}, and Capsule-Forensics-v2~\cite{nguyen1910use}) on four of the proposed Deepfake annotation datasets with respect to 31 demographic and non-demographic attributes\footnote{For experiments, we neglected attributes that are not frequently occurring to avoid wrong conclusions caused by limited testing data. Details can be found in the Appendix and \Cref{tab:attributes_removal}}. 

The results indicate that the investigated datasets are highly imbalanced leading to highly biased detection backbone models when trained on such databases for a large variety of demographic and non-demographic attributes.
 
The observed bias in the detection backbone models can further explain the low generalisability of current Deepfake detectors \cite{DBLP:conf/cvpr/CozzolinoPNV23, DBLP:conf/wacv/XuRP22} across different attributes.
Interestingly, the effect of the imbalanced attributes often differs in detection performance if the attribute is observed on a pristine (nonfake) image or a Deepfake.
The results indicate that the detection backbone models learn several questionable factors that require a deeper investigation. For example, a person smiling or wearing a hat is strongly detected as a \re{real person} despite being a \re{Deepfake} image.
Depending on the application, these factors can lead to biases and, subsequently, strong fairness issues when a \re{fake} video of a smiling woman is detected as \re{a real one}. Conversely, a biased detection backbone model deciding a manipulated video as an unaltered video may lead to security implications. A complete list of such findings from our work \re{along with} the recommendations for future work is provided in Section \ref{sec:FindingsAndFutureWork}.

\section{Related Work}

\subsection{Deepfake Detection}
\label{sec:RelatedWork}
There are two main approaches that are used to detect manipulated media. One focuses on the spatial features extracted from frames of a video. The other utilises temporal features among frames to capture falsified clues.
\begin{itemize}
\item \textbf{Spatial features:} 
Most of the early efforts to detect Deepfake have been made using spatial features extracted from video frames. Researchers have been working on detecting artifacts using unnatural facial features~\cite{li2018ictu}, blending traces~\cite{li2020face}, CNN-generated/GAN-generated fingerprints~\cite{guarnera2020deepfake, chai2020makes}. Some studies have also been conducted in the frequency domain in order to detect artificial image contents~\cite{durall2019unmasking, qian2020thinking}.
\item \textbf{Temporal features:} 
Instead of individual frames, temporal features across frames have also been used recently, for example, unsynchronised color~\cite{mccloskey2019detecting, li2020identification},  and phoney heartbeats 
appearing on faces~\cite{ciftci2020fakecatcher, qi2020deeprhythm} and inconsistent facial information~\cite{yang2019exposing,haliassos2021lips, zheng2021exploring, cozzolino2021id}.
\end{itemize}

Most works have focused on developing Deepfake detection solutions tailored to available datasets. However, these solutions can be imbalanced, leading to bias and low generalisability across different demographic factors. We analyse four Deepfake detection approaches to demonstrate the biased performances for different demographic factors.

\subsection{Deepfake Datasets}
\label{sec:DeepfakeDatasets}
\Cref{tab:DatasetComparison} shows seven popular Deepfake datasets that are popularly used for the development and evaluation of reliable Deepfake detection backbone models. DeepfakeTIMIT~\cite{korshunov2018deepfakes} and FFW~\cite{khodabakhsh2018fake} were published in 2018, followed by DFD~\cite{ffdfd} and FF++~\cite{Rossler_2019_ICCV, li2019faceshifter} in 2019. DFDC~\cite{dolhansky2020deepfake}, Celeb-DF~\cite{li2020celeb} and DF-1.0~\cite{jiang2020deeperforensics1} were released in 2020. Over the years, the size of the datasets has increased in terms of manipulations and the total number of images/videos. However, there are limited efforts to create more balanced datasets for gender and ethnicity. Both Celeb-DF and DF-1.0 maintain parity between males and females. Celeb-DF has a more extensive range of ages, while DF-1.0 holds balanced skin types. Despite these efforts, only a few databases provide additional annotations that could be utilised for developing  Deepfake detection algorithms or testing these for influences of demographic factors. 
In contrast to previous works, we provide high-quality annotations for five popular databases for 47 demographic and non-demographic attributes.
We hope to enable the development and evaluation of balanced and less-biased Deepfake detectors.

\begin{table}[h]
\setlength{\tabcolsep}{7pt}
\renewcommand{\arraystretch}{1.2}
\centering
\caption{\textbf{Comparison of previous bias investigations in Deepfake detection} - this work provides a more comprehensive bias analysis involving more datasets up to 4 and more investigated attributes reaching the quantity of 47.
}
\label{related-work-comparison}
\begin{tabular}{lccc}
\toprule
                                                    & Attributes & Models & Datasets \\ \hline
Hazirbas \textit{et al.}~\cite{hazirbas2021towards} & 4         & 2      & 1        \\
Loc and Yan~\cite{trinh2021examination}             & 2         & \textbf{3}      & 1        \\
Pu \textit{et al.}~\cite{pu2022fairness}             & 2         & 1      & 1     \\
\hline
\textbf{This work}           & \textbf{47}        & \textbf{3}     & \textbf{4}        \\
\bottomrule
\end{tabular}
\end{table}

\begin{table*}[h]
\setlength{\tabcolsep}{7pt}
\renewcommand{\arraystretch}{1.2}
\centering
\caption{\textbf{Overview of popular Deepfake datasets and the proposed annotations databases} - While previous databases lack diverse annotations, the five proposed annotation databases close this gap and provide the resources needed to comprehensively analyse and mitigate bias in Deepfake detection backbone models.}
\label{tab:DatasetComparison}
\footnotesize
\begin{tabular}{llrrrrrr}
\toprule
 & & & \multicolumn{2}{c}{Number of videos} & \multicolumn{2}{c}{Number of frames}  & Annotated\\
  \cmidrule(rl){4-5} \cmidrule(rl){6-7} 
               & Dataset                                                                  & Identities & Pristine & Forged & Pristine & Forged &  Attributes \\
               \hline
\multirow{7}{*}{\rotatebox[origin=c]{90}{Previous works}}
               & DeepfakeTIMIT~\cite{korshunov2018deepfakes}                              & 32         & 320       & 620        & 34.0k     & 68.0k     & -                 \\
               & FFW~\cite{khodabakhsh2018fake}                                           & 150        & -         & 150        & -         & 53k       & -                \\
               & DeepFakeDetection (DFD)~\cite{ffdfd}                                     & 28         & 363       & 3,068      & 315.4k    & 2.2M      & -                 \\
               & FaceForensics++ (FF++)~\cite{Rossler_2019_ICCV, li2019faceshifter}       & 1000       & 1,000     & 5,000      & 300k      & 1.5M      & -                \\
               & Deepfake Detection Challenge Dataset (DFDC)~\cite{dolhansky2020deepfake} & 960        & 23,654    & 104,500    & 7M        & 31M       & -                \\
               & Celeb-DF~\cite{li2020celeb}                                              & 59         & 590       & 5639       & 225.4k    & 2.1M      & -                \\
               & DeeperForensics-1.0 (DF-1.0)~\cite{jiang2020deeperforensics1}                     & 100        & 50,000    & 10,000     & 2.9M      & 14.7M     & 1               \\
               & KoDF~\cite{kwon2021kodf}                                                 & 403        & 62,166    & 175,776    & 135M      & 65.9M     & 2              \\
               \hline
\multirow{5}{*}{\rotatebox[origin=c]{90}{\textbf{This work}}} 
               & \textbf{A-DFD}                                                                    & 28         & 363       & 3068       & 10.8k     & 89.6k     &  \textbf{47}            \\
               & \textbf{A-FF++}                                                                   & 1000       & 1000      & 5000       & 29.8k     & 149.1k    &  \textbf{47}           \\
               & \textbf{A-DFDC}                                                                   & 960        & 23,654    & 104,500    & 54.5k     & 52.6k     &  \textbf{47}        \\
               & \textbf{A-Celeb-DF}                                                               & 59         & 590       & 5639       & 26.3k     & 166.5k    &  \textbf{47}            \\
               & \textbf{A-DF-1.0}                                                    & 100        & 50000     & 10000      & 870.3k    & 321.5k    &  \textbf{47}          \\
               \bottomrule
\end{tabular}
\end{table*}

\subsection{Analysing Bias in Deepfake Detection}
\label{sec:RelatedWorkAnalysingBias}



Internal representations of neural network models preserve attribute-related information of the training data even if it is not directly needed for the model objective \cite{DBLP:conf/icb/TerhorstFDKK20, DBLP:journals/tbbis/TerhorstFDKK21}.
These encoded attribute patterns are reported to lead to biased performance in AI models \cite{Terhorst_2022_FaceBias}.
Although there are many works on studying fairness in AI~\cite{mehrabi2021survey, du2020fairness, Terhorst_2022_FaceBias}, only a few works analyse biases in the Deepfake detection field. Hazirbas \textit{et al.}~\cite{hazirbas2021towards} measured the robustness of Deepfake detection backbone models across four primary dimensions: age, gender, apparent skin type, and lighting. 
They analysed the top five winners of the Deepfake Detection Challenge~\cite{dfdc1, dfdc2, dfdc3, dfdc4, dfdc5} for these attributes and concluded that all methods are biased towards lighter skin tones and fail in subjects with darker skin. 
Loc and Yan~\cite{trinh2021examination} measured the predictive performance of popular Deepfake detectors, MesoInception-4~\cite{afchar2018mesonet}, Xception~\cite{chollet2017xception} and Face X-Ray~\cite{li2020face} on racially balanced datasets for gender and race. Significant disparities were found in predictive performances across races and large representation bias in widely used FF++~\cite{Rossler_2019_ICCV}.
Pu \textit{et al.}~\cite{pu2022fairness} used a subset of the Face2Face dataset in FF++ and investigated MesoInception-4 to verify the existence of gender bias.
Studying bias in Deepfake detection so far is limited to a few demographic factors such as gender and ethnicity. 
In contrast, this work analyses bias of three state-of-the-art Deepfake detection methods on four widely-used Deepfake datasets considering 31 demographic and non-demographic attributes as shown in \Cref{related-work-comparison}. With this work, we provide up to 47 attribute annotations on 4 popular Deepfake datasets.
This work makes it possible to study the bias problem in a more comprehensive and reliable manner.

\section{Methodology}
\label{sec:Methodology}
To analyse different attributes, we first create large-scale annotations of 47 demographic and non-demographic attributes for five Deepfake detection databases. Following this, we conduct a comprehensive bias analysis of the state-of-the-art Deepfake detection methods on these annotated databases.
In the following section, the process for creating the large-scale annotations is described, and methodology for measuring bias is presented.

\subsection{Annotating Deepfake Databases}
\label{AnnotatingDeepfakeDatabases}

\re{We utilize MAAD-Face classifier~\cite{terhorst2021maad}  trained on LFW~\cite{huang2008labeled} and CelebA~\cite{liu2015deep} as source databases to implement a novel annotation-transfer technique that transfers the attribute annotations from several source databases to target databases.}
\re{This approach prioritizes annotations with high confidence predictions, thereby enhancing annotation correctness and minimizing potential biases.}
We annotate five current Deepfake detection databases (DFD~\cite{ffdfd}, FF++~\cite{Rossler_2019_ICCV}, DFDC~\cite{dolhansky2020deepfake}, Celeb-DF~\cite{li2020celeb}, and DF-1.0~\cite{jiang2020deeperforensics1}) in this work.

In the annotation process, each image is assigned with one of three possible labels for an attribute, positive (1), negative (-1), or undefined (0).
A positive annotation for attribute $a$ of an image means that the face in the image has attribute $a$. 
For instance, a face with a positive annotation for 'Young' represents a face of an young individual. 
In contrast, a negative annotation for attribute $a$ of an image means that the face in the image does not possess attribute $a$. 
e further enforce a confidence-driven threshold to assert if an attribute cannot be classified. The confidence score is calculated based on the reliability measure from \cite{DBLP:conf/btas/TerhorstHKZDKK19} and aims at preventing error-prone annotations. Specifically, if the classifier produces a decision for an attribute with a confidence below $90\%$, we annotate the attribute as undefined (0). 
We apply this methodology on five Deepfake detection datasets (DFD, FF++, DFDC, Celeb-DF, DF-1.0), resulting in the annotation datasets A-DFD (4.7M labels), A-FF++ (8.5M labels), A-DFDC (4.6M labels), A-Celeb-DF (9.2M labels) and A-DF-1.0 (38.3M labels), shown in \Cref{tab:DatasetComparison}.
These provide annotations for 47 attributes including information on demographics, skin, hair, beard, face geometry, mouth, nose, and accessories.

\rere{\subsection{Evaluating Bias in Imbalanced Data}}
\label{sec:ControlGroups}

In this study, we assess the bias of a detection backbone model to an attribute $a$ by comparing its performance when the attribute is present versus absent. However, there is a potential issue of an unbalanced distribution of positive and negative labelled testing samples during the experiments. To avoid inaccurate results caused by this imbalance, we introduce a corrected performance measure using control groups of positive and negative samples. We adopt the methods of creating two control groups for each attribute $a$ by randomly selecting $N$ samples from the testing data from \cite{Terhorst_2022_FaceBias}, where $N$ is the number of samples with or without attribute $a$. By doing so, we ensure that each control group has the same number of samples as their counterparts in the real data, thus making the positive and negative control groups independent of individual sample properties and drawn from the same distribution.

Comparing the classification performance of the positive and negative control groups for an attribute $a$ allows us to measure the effect of data imbalance on performance.
If the performances of the negative and positive control groups are similar, the distribution of testing data does not significantly impact the performance.
Contrarily, if the performances of the negative and positive control groups are dissimilar, the unbalanced testing data affects the classification performance.
To measure the bias effect of an attribute $a$ on the performance, we adopt the \textit{relative performance (RP)} measure from \cite{Terhorst_2022_FaceBias}
\begin{align}
    RP_{type}(a) = 1 - \dfrac{err_{type}^{(+)}(a)}{err_{type}^{(-)}(a)},
\end{align}
with $type=\{data, control\}$.
$RP_{type}(a)$ measures the performance differences for an attribute $a$ based on the error rates for a positive $err_{type}^{(+)}(a)$ and a negative $err_{type}^{(-)}(a)$ group.
If the error rates are the same, $RP(a)=0$ and, thus, attribute $a$ does not affect performance.
Positive $RP$ values refer to lower error rates for the positive class (samples with this attribute).
Contrarily, negative $RP$ values refer to lower error rates for the negative class.

To correct this bias in the relative performance measure originating from the unbalanced testing data, we propose the \textit{corrected relative performance} (CRP)
\begin{align}
    CRP(a) &= RP_{data}(a) - RP_{control}(a) 
\end{align}
which describes the difference between the relative performance of the real data $RP_{data}$ and the relative performance of the control groups $RP_{control}$.
The $CRP$ measure simplifies to
\begin{align}
    CRP(a) = \dfrac{err_{control}^{(+)}(a)}{err_{control}^{(-)}(a)} - \dfrac{err_{data}^{(+)}(a)}{err_{data}^{(-)}(a)},
\end{align}
and aims at removing the influence of the testing data distribution from the performance measure.
If biased performance comes only from unbalanced test data, $RP_{data}$ and $RP_{control}$ will be equal, and thus the corrected relative performance $CRP$ will be zero. 
We use the $CRP(a)$ to measure the influence of the presence of attribute $a$ on the performance and thus, to measure bias independently of the testing data parity. 

\begin{figure*}
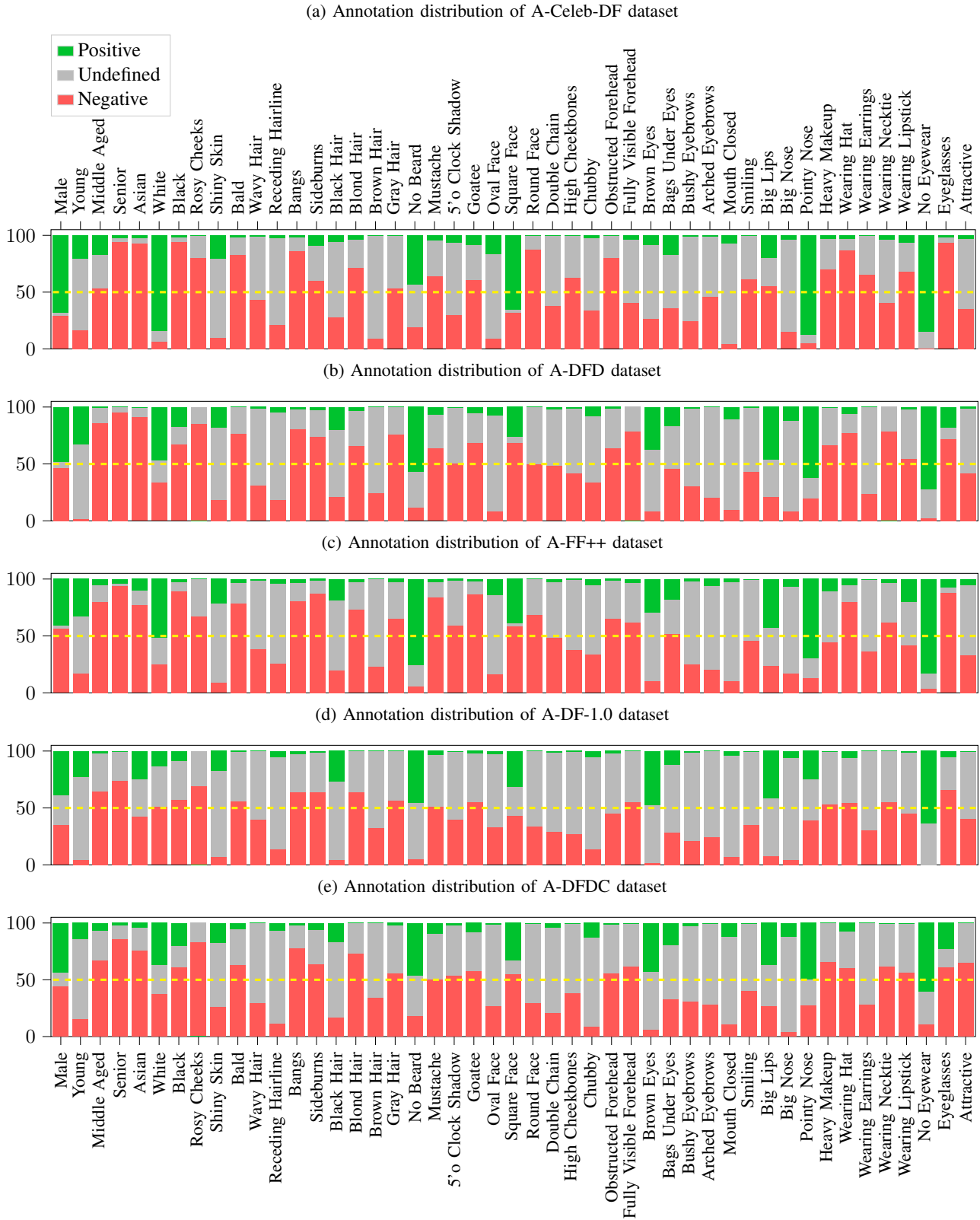

\centering
\caption{
\textbf{Annotation distribution of the annotated Deepfake detection datasets} - The distributions of the proposed dataset annotations are shown with the y-axis presenting percentage. For each  attribute, green indicates the percentage of positive annotations, red indicates the percentage of negatively annotations, and grey represents the percentage of images that have an undefined annotation for the attribute. The distributions show that these databases are highly unbalanced concerning these attributes.}
\label{fig:attributescount}
\subfloat[Annotation distribution of A-Celeb-DF dataset\label{fig:celebcount}]{%
       \resizebox{0.95\linewidth}{!}{\input{imgs/po_ne_count/celeb}}}\\
  \subfloat[Annotation distribution of A-DFD dataset\label{fig:dfdcount}]{%
        \resizebox{0.95\linewidth}{!}{\input{imgs/po_ne_count/dfd}}}\\
  \subfloat[Annotation distribution of A-FF++ dataset\label{fig:ffcount}]{%
        \resizebox{0.95\linewidth}{!}{\input{imgs/po_ne_count/ff}}}\\
  \subfloat[Annotation distribution of A-DF-1.0 dataset\label{fig:deepercount}]{%
        \resizebox{0.95\linewidth}{!}{\input{imgs/po_ne_count/deeper}}}\\
  \subfloat[Annotation distribution of A-DFDC dataset\label{fig:dfdccount}]{%
        \resizebox{0.95\linewidth}{!}{\input{imgs/po_ne_count/dfdc}}}
\end{figure*}

\section{Experimental Setup}

\subsection{Database and Considered Attributes}
\label{DatabaseAndConsideredAttributes}
For the experiments, we choose five widely-used Deepfake detection datasets, DFD~\cite{ffdfd}, FF++~\cite{Rossler_2019_ICCV, li2019faceshifter}, DFDC~\cite{dolhansky2020deepfake}, Celeb-DF~\cite{li2020celeb} and DF-1.0~\cite{jiang2020deeperforensics1}.
Details for the different databases are provided in Table \ref{tab:DatasetComparison}. 30 frames are extracted from the first 300 frames of each video using a 10-frame interval. The faces are detected and aligned using MTCNN \cite{zhang2016joint} for each of the frames. To ensure that enough data is available for analysing bias originating from specific attributes, we ignore attributes where a minimum of 100 positive or negative labelled images are unavailable. Out of the 47 attributes available in the annotated databases, only 31 were included in the bias analysis due to such a curation process. The specific details of this process can be found in Appendix \Cref{tab:attributes_removal}.

\subsection{Deepfake Detection BackBone Models}
\label{DeepfakeDetectionModels}
For the experiments, we choose three well used Deepfake detection backbone models, EfficientNetB0~\cite{tan2019efficientnet}, Xception~\cite{chollet2017xception}, and Capsule-Forensics-v2~\cite{nguyen1910use}. These three networks have been used frequently as backbone networks in the Deepfake detection  \cite{zhao2021multi, li2020face, xu2023learning, luo2021generalizing, montserrat2020Deepfakes, qian2020thinking}. Therefore, we consider it reasonable to use them for the bias analysis.
Furthermore, we have trained and evaluated the three backbone networks with a subject-exclusive train/dev/test for all the attributes. Due to the lack of a standardised protocol for all datasets, we spilt the datasets with a $60\%/20\%/20\%$ proportion for train/val/test respectively.

\begin{itemize}
    \item \textbf{Xception} uses depth-wise separable convolutions to reduce the computational cost of traditional convolutions while maintaining high accuracy. This is achieved by performing spatial and channel convolutions separately, allowing for more efficient image feature processing. It is a highly effective deep learning architecture for image recognition tasks that require high accuracy and computational efficiency.
    \item \textbf{EfficientNet} is a model scaling method that uses a simple yet highly effective compound coefficient to scale up CNNs in a more structured manner, balancing the network's depth, width, and resolution to optimize its performance on a given resource budget. The architecture includes several novel features, including a mobile inverted bottleneck block, squeeze-and-excitation optimisation, and stochastic depth regularisation, further improving its performance. In our paper, we select the most lightweight version of EfficientNet, EfficientNetB0, to showcase its effectiveness.
    \item \textbf{Capsule-Forensics-v2} uses capsules to extract facial features and their spatial relationships from the input image to detect discrepancies. It incorporates a novel loss function that encourages disentangled representations, improving forgery detection accuracy. The model has demonstrated high effectiveness in detecting image manipulations, including copy-move, splicing, and face morphing.
\end{itemize}

\subsection{Evaluation Metrics}
\label{sec:EvaluationMetrics}
Previous work on Deepfake detection has reported its results based on a simple accuracy measure \cite{trinh2021examination, pu2022fairness}. However, dealing with unbalanced testing data is the norm, and a simple accuracy measure is vulnerable to this. We further notice many attributes being unbalanced in terms of the positive/negative labels from \Cref{fig:attributescount}. We, therefore, make use of a balanced accuracy measure, which computes the arithmetic mean of the sensitivity and specificity and is more robust to unbalanced data~\cite{brodersen2010balanced}. More precisely, we report the performances in terms of error rates (1-balanced accuracy) since this work investigates bias issues driven by inaccurate predictions.

\begin{figure*}[htp]
\centering
  \subfloat[A-Celeb-DF\label{fig:celebpearson}]{%
       \resizebox{.34\linewidth}{!}{
\begin{tikzpicture}

\begin{axis}[
width=6cm,
height=6cm,
colorbar,
colorbar style={width=9, ytick={-1.0,-0.5,0,0.5,1.0},yticklabels={
  \ensuremath{-}1.0,
  \ensuremath{-}0.5,
  0.0,
  0.5,
  1.0
},ylabel={Correlation coefficient}},
colormap={mymap}{[1pt]
  rgb(0pt)=(0.647058823529412,0,0.149019607843137);
  rgb(1pt)=(0.843137254901961,0.188235294117647,0.152941176470588);
  rgb(2pt)=(0.956862745098039,0.427450980392157,0.262745098039216);
  rgb(3pt)=(0.992156862745098,0.682352941176471,0.380392156862745);
  rgb(4pt)=(0.996078431372549,0.87843137254902,0.545098039215686);
  rgb(5pt)=(1,1,0.749019607843137);
  rgb(6pt)=(0.850980392156863,0.937254901960784,0.545098039215686);
  rgb(7pt)=(0.650980392156863,0.850980392156863,0.415686274509804);
  rgb(8pt)=(0.4,0.741176470588235,0.388235294117647);
  rgb(9pt)=(0.101960784313725,0.596078431372549,0.313725490196078);
  rgb(10pt)=(0,0.407843137254902,0.215686274509804)
},
point meta max=1,
point meta min=-1,
tick align=outside,
tick pos=left,
x grid style={white!69.0196078431373!black},
yticklabel style={font=\large},
xmin=0, xmax=10,
xtick style={color=black},
xtick={0.5,1.5,2.5,3.5,4.5,5.5,6.5,7.5,8.5,9.5},
xticklabel style={rotate=90.0, font=\large},
xtick pos=top,
xticklabels={
  Mustache,
  Goatee,
  Square Face,
  Double Chain,
  Chubby,
  Fully Visible Forehead,
  Big Lips,
  Pointy Nose,
  Heavy Makeup,
  Wearing Lipstick
},
y dir=reverse,
y grid style={white!69.0196078431373!black},
ymin=0, ymax=10,
ytick style={color=black},
yticklabel style={font=\large},
ytick={0.5,1.5,2.5,3.5,4.5,5.5,6.5,7.5,8.5,9.5},
yticklabels={
  Mustache,
  Goatee,
  Square Face,
  Double Chain,
  Chubby,
  Fully Visible Forehead,
  Big Lips,
  Pointy Nose,
  Heavy Makeup,
  Wearing Lipstick
}
]
\addplot graphics [includegraphics cmd=\pgfimage,xmin=0, xmax=10, ymin=10, ymax=0] {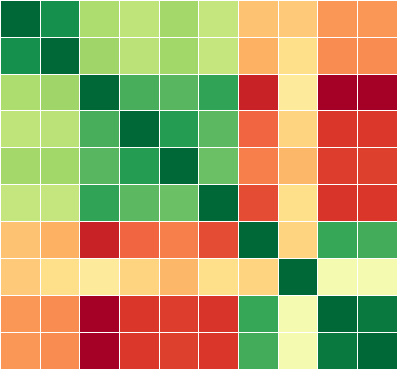};

\end{axis}

\end{tikzpicture}}}
  \subfloat[A-DFD\label{fig:dfdpearson}]{%
        \resizebox{.32\linewidth}{!}{
\begin{tikzpicture}

\begin{axis}[
width=6cm,
height=6cm,
colorbar,
colorbar style={width=9, ytick={-1.0,-0.5,0,0.5,1.0},yticklabels={
  \ensuremath{-}1.0,
  \ensuremath{-}0.5,
  0.0,
  0.5,
  1.0
},ylabel={Correlation coefficient}},
colormap={mymap}{[1pt]
  rgb(0pt)=(0.647058823529412,0,0.149019607843137);
  rgb(1pt)=(0.843137254901961,0.188235294117647,0.152941176470588);
  rgb(2pt)=(0.956862745098039,0.427450980392157,0.262745098039216);
  rgb(3pt)=(0.992156862745098,0.682352941176471,0.380392156862745);
  rgb(4pt)=(0.996078431372549,0.87843137254902,0.545098039215686);
  rgb(5pt)=(1,1,0.749019607843137);
  rgb(6pt)=(0.850980392156863,0.937254901960784,0.545098039215686);
  rgb(7pt)=(0.650980392156863,0.850980392156863,0.415686274509804);
  rgb(8pt)=(0.4,0.741176470588235,0.388235294117647);
  rgb(9pt)=(0.101960784313725,0.596078431372549,0.313725490196078);
  rgb(10pt)=(0,0.407843137254902,0.215686274509804)
},
point meta max=1,
point meta min=-1,
tick align=outside,
tick pos=left,
x grid style={white!69.0196078431373!black},
xmin=0, xmax=13,
xtick style={color=black},
xtick={0.5,1.5,2.5,3.5,4.5,5.5,6.5,7.5,8.5,9.5,10.5,11.5,12.5},
xticklabel style={rotate=90.0, font=\large},
xtick pos=top,
xticklabels={
  Black,
  Black Hair,
  Blond Hair,
  No Beard,
  Mustache,
  Goatee,
  Square Face,
  Chubby,
  Big Lips,
  PointyNose,
  Heavy Makeup,
  Wearing Lipstick,
  Attractive
},
y dir=reverse,
y grid style={white!69.0196078431373!black},
ymin=0, ymax=13,
ytick style={color=black},
ytick={0.5,1.5,2.5,3.5,4.5,5.5,6.5,7.5,8.5,9.5,10.5,11.5,12.5},
yticklabel style={font=\large},
yticklabels={
  Black,
  Black Hair,
  Blond Hair,
  No Beard,
  Mustache,
  Goatee,
  Square Face,
  Chubby,
  Big Lips,
  PointyNose,
  Heavy Makeup,
  Wearing Lipstick,
  Attractive
}
]
\addplot graphics [includegraphics cmd=\pgfimage,xmin=0, xmax=13, ymin=13, ymax=0] {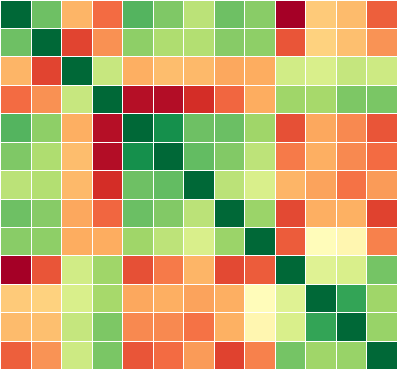};
\end{axis}


\end{tikzpicture}}}
  \subfloat[A-FF++\label{fig:ffpearson}]{%
        \resizebox{.33\linewidth}{!}{
\begin{tikzpicture}

\begin{axis}[
width=6cm,
height=6cm,
colorbar,
colorbar style={width=9, ytick={-1.0,-0.5,0,0.5,1.0},yticklabels={
  \ensuremath{-}1.0,
  \ensuremath{-}0.5,
  0.0,
  0.5,
  1.0
},ylabel={Correlation coefficient}},
colormap={mymap}{[1pt]
  rgb(0pt)=(0.647058823529412,0,0.149019607843137);
  rgb(1pt)=(0.843137254901961,0.188235294117647,0.152941176470588);
  rgb(2pt)=(0.956862745098039,0.427450980392157,0.262745098039216);
  rgb(3pt)=(0.992156862745098,0.682352941176471,0.380392156862745);
  rgb(4pt)=(0.996078431372549,0.87843137254902,0.545098039215686);
  rgb(5pt)=(1,1,0.749019607843137);
  rgb(6pt)=(0.850980392156863,0.937254901960784,0.545098039215686);
  rgb(7pt)=(0.650980392156863,0.850980392156863,0.415686274509804);
  rgb(8pt)=(0.4,0.741176470588235,0.388235294117647);
  rgb(9pt)=(0.101960784313725,0.596078431372549,0.313725490196078);
  rgb(10pt)=(0,0.407843137254902,0.215686274509804)
},
point meta max=1,
point meta min=-1,
tick align=outside,
tick pos=left,
x grid style={white!69.0196078431373!black},
xmin=0, xmax=11,
xtick style={color=black},
xtick={0.5,1.5,2.5,3.5,4.5,5.5,6.5,7.5,8.5,9.5,10.5},
xticklabel style={rotate=90.0, font=\large},
xtick pos=top,
xticklabels={
  Wavy Hair,
  Mustache,
  Goatee,
  Square Face,
  Double Chain,
  Chubby,
  Fully Visible Forehead,
  Pointy Nose,
  Heavy Makeup,
  Wearing Lipstick,
  Attractive
},
y dir=reverse,
y grid style={white!69.0196078431373!black},
ymin=0, ymax=11,
ytick style={color=black},
yticklabel style={font=\large},
ytick={0.5,1.5,2.5,3.5,4.5,5.5,6.5,7.5,8.5,9.5,10.5},
yticklabels={
  Wavy Hair,
  Mustache,
  Goatee,
  Square Face,
  Double Chain,
  Chubby,
  Fully Visible Forehead,
  Pointy Nose,
  Heavy Makeup,
  Wearing Lipstick,
  Attractive
}
]
\addplot graphics [includegraphics cmd=\pgfimage,xmin=0, xmax=11, ymin=11, ymax=0] {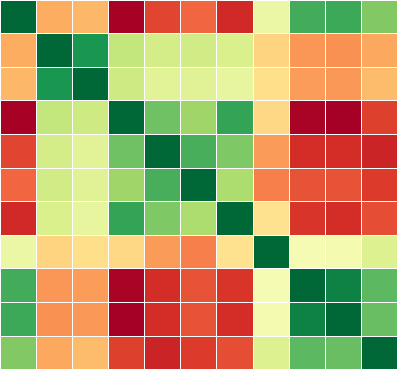};
\end{axis}

\end{tikzpicture}}}\\
  \subfloat[A-DF-1.0\label{fig:deeperpearson}]{%
        \resizebox{.34\linewidth}{!}{
\begin{tikzpicture}

\begin{axis}[
width=6cm,
height=6cm,
colorbar,
colorbar style={width=9, ytick={-1.0,-0.5,0,0.5,1.0},yticklabels={
  \ensuremath{-}1.0,
  \ensuremath{-}0.5,
  0.0,
  0.5,
  1.0
},ylabel={Correlation coefficient}},
colormap={mymap}{[1pt]
  rgb(0pt)=(0.647058823529412,0,0.149019607843137);
  rgb(1pt)=(0.843137254901961,0.188235294117647,0.152941176470588);
  rgb(2pt)=(0.956862745098039,0.427450980392157,0.262745098039216);
  rgb(3pt)=(0.992156862745098,0.682352941176471,0.380392156862745);
  rgb(4pt)=(0.996078431372549,0.87843137254902,0.545098039215686);
  rgb(5pt)=(1,1,0.749019607843137);
  rgb(6pt)=(0.850980392156863,0.937254901960784,0.545098039215686);
  rgb(7pt)=(0.650980392156863,0.850980392156863,0.415686274509804);
  rgb(8pt)=(0.4,0.741176470588235,0.388235294117647);
  rgb(9pt)=(0.101960784313725,0.596078431372549,0.313725490196078);
  rgb(10pt)=(0,0.407843137254902,0.215686274509804)
},
point meta max=1,
point meta min=-1,
tick align=outside,
tick pos=left,
x grid style={white!69.0196078431373!black},
xmin=0, xmax=13,
xtick style={color=black},
xtick={0.5,1.5,2.5,3.5,4.5,5.5,6.5,7.5,8.5,9.5,10.5,11.5,12.5},
xticklabel style={rotate=90.0, font=\large},
xtick pos=top,
xticklabels={
  Bald,
  Mustache,
  Goatee,
  Square Face,
  Double Chain,
  Chubby,
  Obstructed Forehead,
  Fully Visible Forehead,
  Pointy Nose,
  Heavy Makeup,
  Wearing Lipstick,
  Eyeglasses,
  Attractive
},
y dir=reverse,
y grid style={white!69.0196078431373!black},
ymin=0, ymax=13,
ytick style={color=black},
yticklabel style={font=\large},
ytick={0.5,1.5,2.5,3.5,4.5,5.5,6.5,7.5,8.5,9.5,10.5,11.5,12.5},
yticklabels={
  Bald,
  Mustache,
  Goatee,
  Square Face,
  Double Chain,
  Chubby,
  Obstructed Forehead,
  Fully Visible Forehead,
  Pointy Nose,
  Heavy Makeup,
  Wearing Lipstick,
  Eyeglasses,
  Attractive
}
]
\addplot graphics [includegraphics cmd=\pgfimage,xmin=0, xmax=13, ymin=13, ymax=0] {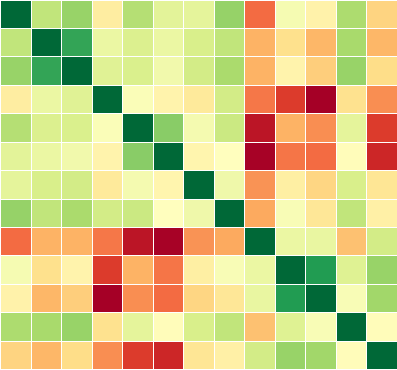};
\end{axis}

\end{tikzpicture}}}
  \subfloat[A-DFDC\label{fig:dfdcpearson}]{%
        \resizebox{.31\linewidth}{!}{
\begin{tikzpicture}

\begin{axis}[
width=6cm,
height=6cm,
colorbar,
colorbar style={width=9, ytick={-1.0,-0.5,0,0.5,1.0},yticklabels={
  \ensuremath{-}1.0,
  \ensuremath{-}0.5,
  0.0,
  0.5,
  1.0
},ylabel={Correlation coefficient}},
colormap={mymap}{[1pt]
  rgb(0pt)=(0.647058823529412,0,0.149019607843137);
  rgb(1pt)=(0.843137254901961,0.188235294117647,0.152941176470588);
  rgb(2pt)=(0.956862745098039,0.427450980392157,0.262745098039216);
  rgb(3pt)=(0.992156862745098,0.682352941176471,0.380392156862745);
  rgb(4pt)=(0.996078431372549,0.87843137254902,0.545098039215686);
  rgb(5pt)=(1,1,0.749019607843137);
  rgb(6pt)=(0.850980392156863,0.937254901960784,0.545098039215686);
  rgb(7pt)=(0.650980392156863,0.850980392156863,0.415686274509804);
  rgb(8pt)=(0.4,0.741176470588235,0.388235294117647);
  rgb(9pt)=(0.101960784313725,0.596078431372549,0.313725490196078);
  rgb(10pt)=(0,0.407843137254902,0.215686274509804)
},
point meta max=1,
point meta min=-1,
tick align=outside,
tick pos=left,
x grid style={white!69.0196078431373!black},
xmin=0, xmax=10,
xtick style={color=black},
xtick={0.5,1.5,2.5,3.5,4.5,5.5,6.5,7.5,8.5,9.5},
xticklabel style={rotate=90.0, font=\large},
xtick pos=top,
xticklabels={
  Black,
  Shiny Skin,
  No Beard,
  Mustache,
  Goatee,
  Square Face,
  Mouth Closed,
  Big Lips,
  Pointy Nose,
  Wearing Lipstick
},
y dir=reverse,
y grid style={white!69.0196078431373!black},
ymin=0, ymax=10,
ytick style={color=black},
yticklabel style={font=\large},
ytick={0.5,1.5,2.5,3.5,4.5,5.5,6.5,7.5,8.5,9.5},
yticklabels={
  Black,
  Shiny Skin,
  No Beard,
  Mustache,
  Goatee,
  Square Face,
  Mouth Closed,
  Big Lips,
  Pointy Nose,
  Wearing Lipstick
}
]
\addplot graphics [includegraphics cmd=\pgfimage,xmin=0, xmax=10, ymin=10, ymax=0] {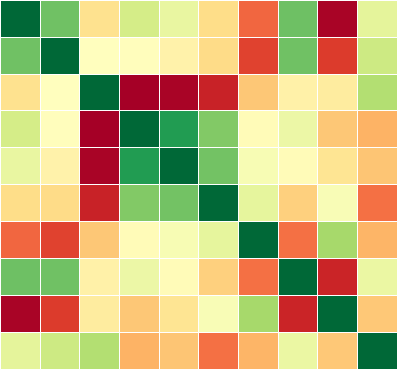};
\end{axis}

\end{tikzpicture}}}
\caption{\textbf{Attribute annotation correlations of the Deepfake detection databases} - The 20 most positive and negative (Pearson) correlations are shown for each of the five databases. Green indicates positive correlations, while red indicates negative correlations. For working with these databases, the highly-correlating attributes should be considered to prevent misinterpretations.
}
\label{fig:pearsoncorrelation}
\end{figure*}

\section{Results}
This section presents our findings on the presence of bias in Deepfake detection datasets utilising our proposed annotations. We analysed the relationship among various variables regarding RP-vs-CRP and PDRP-vs-DDRP and used 24 plots to visualize these relationships. As we discuss these findings, it is crucial to keep in mind the concepts of causality and correlation in research and statistics. Causality refers to the relationship between cause and effect, while correlation measures the degree to which two or more variables are associated. It is important to note that correlation does not necessarily imply causality. Therefore, our results will focus on explaining correlation, and we will leave an in-depth exploration of causality for future studies.

\subsection{Analysing Database Annotations}
\label{sec:AnalysingDatabaseAnnotations}
\subsubsection{Attribute Statistics}
\Cref{fig:attributescount} shows the annotation distribution of the five annotated Deepfake detection datasets. For each attribute, green indicates the percentage of positive annotations, red indicates the percentage of negative annotations, and grey represents the percentage of images that have an undefined annotation for the attribute. According to the data given by Celeb-DF~\cite{li2020celeb}, this dataset contains male of 56.8\% and female 43.2\%, and in Figure \ref{fig:celebcount}, the percentage of male (positive) is 70.15\% and the percentage of female (negative) is 29.85\%. The reason of the increased gap between gender asymmetry is that Celeb-DF only generates Deepfake videos using the same gender, so the number of differences between male and female are enlarged among the synthesised videos. We notice that most databases are quite balanced for the gender attribute.

However, there is a big imbalance with respect to skin colour, especially in Celeb-DF where people with white skin tones occupy the vast majority of this dataset.
The big gap between numbers corresponds to the number disparity of Celeb-DF~\cite{li2020celeb} (5.1\% Asian, 6.8\% African, and 88.1\% Caucasian) which clarifies the high accuracy of the MAAD-classifier. 
The DFDC dataset appears to have a noticeable under-representation of individuals of Asian descent. Furthermore, there is a prevalence of white individuals in both the DFD and FF++ datasets.
The variations in skin color distribution across different datasets may result in biases in the Deepfake detection system.

\textit{To conclude, it is clearly visible that the investigated Deepfake detection databases (DFD, FF++, DF-1.0, and DFDC) are strongly imbalanced with respect to most analysed attributes.
Future work should consider creating balanced datasets to prevent any potential biases in Deepfake detection algorithms when such datasets are used for training.}

\subsubsection{Attribute Correlations}
\label{sec:AttributeCorrelation}

We present 20 most positive and negative pairwise attribute correlations in \Cref{fig:pearsoncorrelation} to understand the quality of the labels and potential biases in the attribute space. For instance, we notice in \Cref{fig:celebpearson} that the attributes of Mustache and Goatee are highly correlating with each other. A high correlation also occurs between Heavy Makeup and Wearing Lipstick. This is easy to understand as the former attributes are mainly associated with males, and the latter ones mainly are with females. 
In contrast, Mustache and Goatee are negatively correlated to Heavy Makeup, and Wearing Lipsticks. 
Similar patterns are observable across different attribute correlations.
The presence of negative correlations, such as the inverse relationship between \textit{No Beard} and \textit{Mustache}, as well as \textit{Goatee}, highlights the quality of the annotations.
It still should be noted that some correlations might also origin from the MAAD-classifier.
Most of these correlations can be explained with background knowledge of the databases.
For instance, the Celeb-DF dataset contains mainly images of celebrities in which these are presenting themselves to the camera. Therefore, a high correlation between Heavy Makeup and Wearing Lipstick is observed which may not necessarily represent real-world Deepfake of non-celebrities.

\textit{
To conclude, our investigation has identified attribute pairs within the databases that exhibit strong correlations. It is imperative that future studies utilising these datasets and annotations consider these attribute correlations to avoid any misinterpretations that may result in biases. By acknowledging and accounting for these correlations, we can enhance the accuracy and fairness of any analysis or application of these databases.
}

\subsubsection{Annotation Correctness}
To evaluate the effectiveness of the proposed annotations, we have adopted \Cref{tab:AnnotationCorrectness} from the MAAD-classifier~\cite{terhorst2021maad}. This table verifies the accuracy of the MAAD-classifier for the attributes utilised in our study.
This table originates from ~\cite{terhorst2021maad} and shows the attribute correctness of the classifier with respect to three human evaluators.
For each attribute, 100 images with and 100 images without this attribute were chosen randomly and shown to the evaluators to determine the true attribute label for each image.
If the evaluators disagreed on an attribute, majority voting was used to decide on a label.
Then, the accuracy, precision, and recall of the classifier predictions are calculated based on the ground truth provided by the human evaluators.
The results are shown in Table \ref{tab:AnnotationCorrectness}.
For most attributes, the classifier agrees with human evaluators, resulting in an average accuracy of 92\%, precision of 90\%, and recall of 94\%.
Compared to similar facial annotation databases, such as LFW~\cite{huang2008labeled} (72\% accuracy, 61\% precision, 84\% recall) and CelebA~\cite{liu2015deep} (85\% accuracy, 83\% precision, 89\% recall)~\cite{terhorst2021maad}, the proposed annotations are of high correctness.

\textit{The annotations provided in this work are of higher quality than the annotations provided for previous databases and we assert them to be suitable for analysing bias in Deepfake detection. Future works can make use of these attributes for developing and analysing bias-mitigating approaches in Deepfake detection.}

\begin{table}[]
\renewcommand{\arraystretch}{1.1}
\setlength{\tabcolsep}{4pt}
\caption{
\textbf{Annotation Correctness Study} - Annotation correctness of the utilized annotation generator is compared with the annotations of three human evaluators\cite{terhorst2021maad}. Compared to similar large-scale facial annotation classifiers used for databases, such as LFW~\cite{huang2008labeled} (72\% annotation accuracy) and CelebA~\cite{liu2015deep} (85\% annotation accuracy), the proposed annotations are of high correctness~\cite{terhorst2021maad} (92\% annotation accuracy).}
\label{tab:AnnotationCorrectness}
\centering
\begin{tabular}{llrrr}
\Xhline{2\arrayrulewidth}
Category      & Attribute              & Accuracy & Precision & Recall \\
\hline
Demographics  & Male                   & 0.99     & 0.98      & 1.00   \\
              & Young                  & 0.99     & 1.00      & 0.98   \\
              & Asian                  & 0.90     & 0.88      & 0.92   \\
              & White                  & 0.89     & 1.00      & 0.82   \\
              & Black                  & 0.94     & 0.90      & 0.98   \\
Skin          & Shiny Skin             & 0.77     & 0.84      & 0.74   \\
Hair          & Bald                   & 0.96     & 0.92      & 1.00   \\
              & Wavy Hair              & 0.99     & 1.00      & 0.98   \\
              & Receding Hairline      & 0.77     & 0.54      & 1.00   \\
              & Bangs                  & 0.98     & 0.96      & 1.00   \\
              & Black Hair             & 0.98     & 0.96      & 1.00   \\
              & Blond Hair             & 1.00     & 1.00      & 1.00   \\
Beard         & No Beard               & 0.98     & 1.00      & 0.96   \\
              & Mustache               & 0.98     & 0.98      & 0.98   \\
              & Goatee                 & 0.95     & 0.90      & 1.00   \\
Face Geometry & Oval Face              & 0.81     & 0.90      & 0.76   \\
              & Square Face            & 0.80     & 0.78      & 0.81   \\
              & Double Chin            & 0.94     & 0.88      & 1.00   \\
              & Chubby                 & 0.94     & 0.88      & 1.00   \\
              & Obstructed Forehead    & 0.91     & 0.94      & 0.89   \\
              & Fully Visible Forehead & 0.80     & 0.75      & 1.00   \\
Mouth         & Mouth Closed           & 0.84     & 0.80      & 0.87   \\
              & Smiling                & 0.95     & 1.00      & 0.91   \\
              & Big Lips               & 0.70     & 0.50      & 0.83   \\
Nose          & Big Nose               & 0.97     & 0.98      & 0.96   \\
              & Pointy Nose            & 0.88     & 0.88      & 0.88   \\
Accessories   & Heavy Makeup           & 0.98     & 0.98      & 0.98   \\
              & Wearing Hat            & 0.92     & 0.84      & 1.00   \\
              & Wearing Lipstick       & 0.95     & 0.90      & 1.00   \\
              & No Eyewear             & 0.98     & 0.98      & 0.98   \\
              & Eyeglasses             & 0.90     & 0.80      & 1.00   \\
Other         & Attractive             & 1.00     & 1.00      & 1.00   \\
\hline
              &                        & 0.92     & 0.90      & 0.94  \\
\Xhline{2\arrayrulewidth}
\end{tabular}
\vspace{-10pt}
\end{table}

\subsection{Analysing Bias in Deepfake Detection}
\label{sec:AnalysingBiasInDeepFakeDetection}

To understand the bias in Deepfake detection, we will first study the general detection performance in presence of several potentially imbalanced attributes and secondly, analysing the detection performance in presence of these attributes separately on pristine and fake data. We exclude DF-1.0 dataset as the detection methods did not produce high enough errors necessary to analyse biased behaviours. The detailed results are shown in Appendices \Cref{tab:Deeperf-effb0}, \Cref{tab:Deeperf-xception}, and \Cref{tab:Deeperf-capnet} due to page limits.

\subsubsection{Investigating General Bias Issues}
\label{sec:CorrectedRelativePerformance}

\begin{figure*}[htp]
    \centering
    \includegraphics[width=\linewidth]{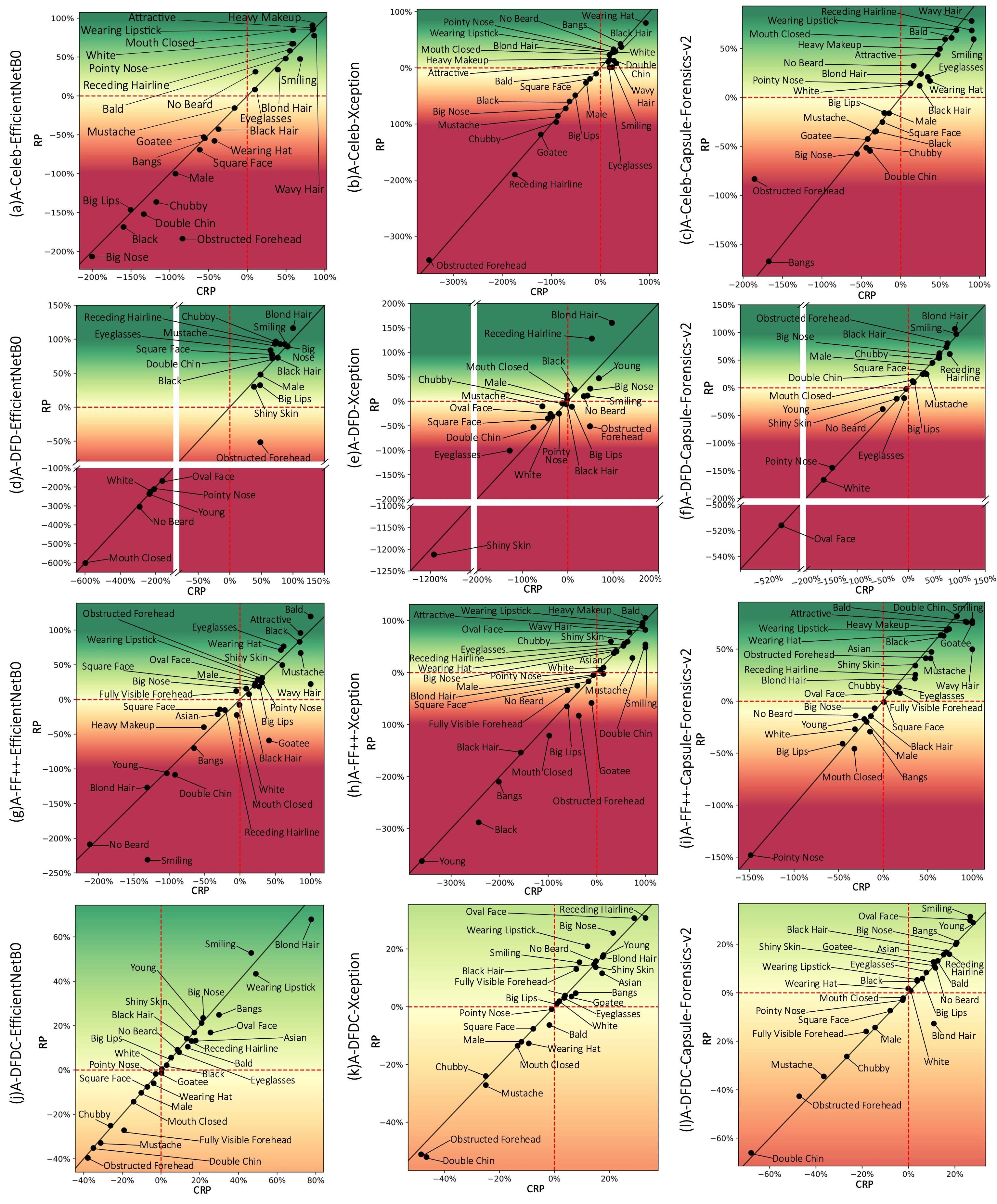}
    \caption{\textbf{Bias analysis} - The relative performance $RP$ is reported with respect to the corrected relative performance $CRP$ using three Deepfake detection backbone models, EfficientNetB0~\cite{tan2019efficientnet}, Xception~\cite{chollet2017xception}, and Capsule-Forensics-v2~\cite{nguyen1910use} on four annotated databases, A-Celeb-DF, A-DFD, A-FF++, and A-DFDC. Many attributes strongly influence the detection performance.} 
    \label{fig:crp-vs-rp}
\end{figure*}

This section analyses the general bias issues in Deepfake detection based on RP-vs-CRP plots as shown in Figure \ref{fig:crp-vs-rp}.
In these plots, the relative performance (RP) for each attribute is shown with respect to the corrected relative performance (CRP). 
As mentioned in Section \ref{sec:ControlGroups}, $RP$ describes the ratio of the performance for images with a certain attribute versus the performance without this attribute. 
Consequently, $RP(a)=-100\%$ for an attribute $a$ means that the error is twice as high if the image has this attribute than without it.
Since the testing data is imbalanced for many attributes, the CRP was introduced in Section \ref{sec:ControlGroups} to remove the influence of data imbalance.
Consequently, attributes that lie in the top area (green) of the RP-vs-CRP plots indicate an increased detection performance and, contrarily, attributes that lie in the bottom (red) indicate increased detection errors.
Moreover, each plot contains a bisectrix line where the attributes close to this line are less affected by imbalanced testing data than attributes away from it.

The RP-vs-CRP plots in Figure \ref{fig:crp-vs-rp} are shown for three models on four Deepfake detection databases.
The plots show strong influences of most of the investigated attributes on the performance, indicating strongly biased Deepfake detectors.
For instance, the analysis of EfficientNetB0 on Celeb-DF shows that having a big nose/big lips/or being black or chubby leads to more than twice the detection errors compared to images without these attributes. 
This shows serious fairness issues of these models when these are applied in real-world applications for specific category of people. In general, most attributes can be observed as strong factors leading to unfair performance differences in DeepFake detection.

Moreover, we have observed that training Deepfake detection backbone models on various datasets results in significant variations in the influence of attributes on detection performance.
For instance, the misclassification of the pattern \textit{Obstructed Forehead} is observed in the Celeb-DF and DFDC datasets.
This finding suggests that both the selection of Deepfake detection backbone networks and the choice of datasets may significantly impact bias in the system.

\textit{To conclude, the experimental results demonstrate that the analysed Deepfake detection backbone models are strongly biased against a variety of demographic and non-demographic attributes. 
The variation of the biased performances across the models and databases indicates that this bias originates from several sources such as unbalanced training data, the utilised network, and their training process.
The observed attribute-related variation in performances shows a strong need for mitigating bias in Deepfake detection models.
}

\begin{figure*}[htp]
    \centering
    \includegraphics[width=\linewidth]{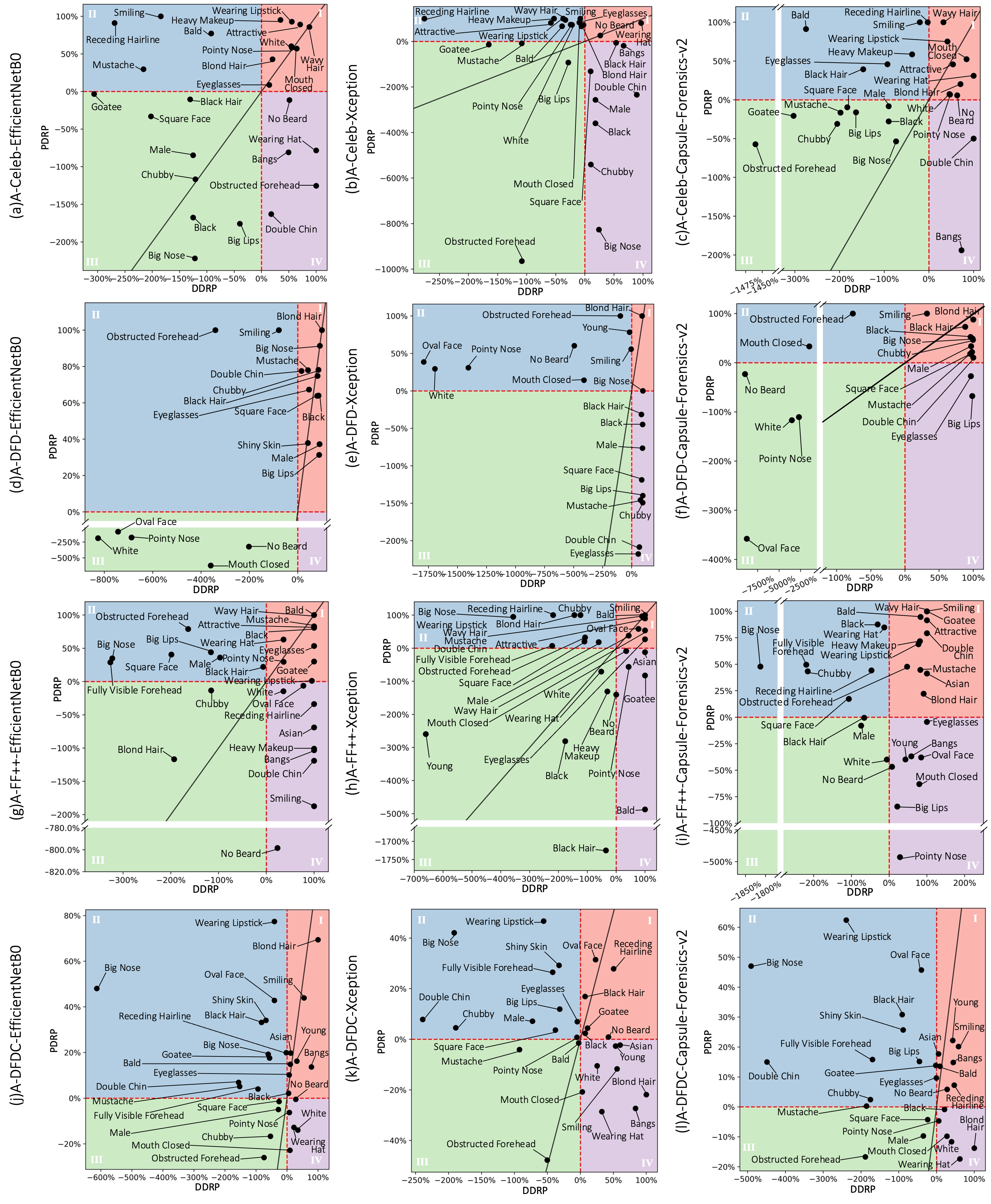}
    \caption{\textbf{Bias analysis on Pristine Data and Deepfake Data} - The $CRP$ on the pristine data (PDRP) is reported with respect to the $CRP$ on Deepfake data (DDRP) using three Deepfake detection backbone models, EfficientNetB0~\cite{tan2019efficientnet}, Xception~\cite{chollet2017xception}, and Capsule-Forensics-v2~\cite{nguyen1910use} on our four annotated databases, A-Celeb-DF, A-DFD, A-FF++, and A-DFDC. Many of the attributes strongly influence the detection performance.}
    \label{fig:real-vs-fake}
\end{figure*}

\subsubsection{Investigating Bias Issues in Pristine and Fake Data}
To investigate the bias issues in DeepFake detection in more detail, we conduct another analysis for pristine and fake data individually in this section.
The results of this analysis is shown in Figure \ref{fig:real-vs-fake}, for three Deepfake detection backbone models on four databases. 
The pristine data relative performance ($PDRP$) refers to the $CRP$ that is only evaluated on pristine data and, analogous, the Deepfake data relative performance ($DDRP$) refers to the $CRP$ that is calculated on fake data only.
For an attribute $a$, a negative $CRP$ on the pristine data means that people having this attribute are more likely to be falsely detected as fakes than people without these attributes.
A negative $CRP$ on fake data means that fake images that are generated with such an attribute are less likely to be detected as fake and thus, demonstrate weak points that attackers are likely to exploit.
Each plot also contains bisectrix line where attributes that lie close to this line have a similar affect on pristine data than on Deepfake.
Attributes placed above this line have a higher $CRP$ on \re{the pristine than on the Deepfake.}
Conversely, attributes below this line have a higher $CRP$ on \re{Deepfake than on the pristine data.}

The results clearly show that the effect of the investigated attributes on the detection performance strongly differ between pristine and fake data since most attributes lie far away from the bisectrix line.
Analysing the four quadrants of the plots shows that in most cases the attributes are distributed in all four.
Attributes in quadrant I (top right), indicate that the attributes have the same positive effect on the performance, while attribute in quadrant III (bottom left), have the same negative effect on the detection performance.
Observed performance in these areas indicates a similar biased effect on the decision.
Attributes in quadrant II (top left) and IV (bottom right) show the opposite effect on the detection performance on pristine and Deepfake data.
Consequently, for attributes in these areas, the model learnt the critical assumptions that the presence of the attribute is an indicator for Deepfake detection decision.
For instance, the analysis of EfficientNetB0 on Celeb-DF for attribute wearing hat, shows a positive \re{$DDPR$} $\approx 100\%$ and a negative \re{$PDPR$} $\approx -75\%$.
Consequently, if a \re{person in a Deepfake image} is wearing a hat the model comes twice as often to the right decision than if the person is not wearing a hat.
Conversely, if a \re{real person} with hat is analysed by the model it leads to nearly twice as many errors as without a hat.
The model in this case has seemingly learnt the presence of a hat as a strong indicator for the \re{Deepfake} data.
Such observations point to questionable assumptions learned by the network and can result in biased detection performance.

In general, the observations reveal similar trends and patterns corresponding to the investigation from Section \ref{sec:CorrectedRelativePerformance}.
The biased performances for the different attributes vary across the utilised models and training databases.
If a \re{Deepfake} person has a goatee, a big nose, is chubby, male, or black, the probability that the model (EfficientNetB0 on Celeb-DF) comes to a wrong decision is doubled compared to persons without these attributes. The detection models therefore show strong biases leading to fairness issues in real-life applications if deployed without considering the attribute distribution.
It should also be kept in mind that this analysis limits its investigation to the influence of single attributes on the detection performance.
The analysis of multiple attributes can be asserted to lead to an exponential increase in its bias effects. However, this aspect is not considered in this work.

\textit{To conclude, the impact of biased performance for the analysed attributes on detection accuracy varies significantly between pristine and fake data for several attributes.
The results suggest that the models learn several questionable assumptions that the presence of a certain attribute, such as if the person is smiling or wears a hat, is an strong indicator for Deepfake detection decision.
Lastly, the investigated Deepfake detection backbone models have demonstrated unfair behavior, with a significant increase in the probability of making incorrect decisions when presented with specific attributes such as having a big nose or belonging to a particular gender or race. This bias in current Deepfake detectors affects their accuracy and limits their generalisability. In other words, these biases may cause the Deepfake detectors to perform well on certain datasets or scenarios, but may fail to perform effectively in others, especially those where such attributes are different or not present. Therefore, addressing these biases and improving the generalisability of Deepfake detectors is crucial to ensure their robustness and reliability in real-world applications.
}

\section{Key Findings \& Recommendations For Future Works}
\label{sec:FindingsAndFutureWork}
In the following, we summarise our key findings from our bias investigation of three Deepfake detection backbone models in four databases with respect to 31 demographic and non-demographic attributes:

\begin{itemize}
\item \textbf{Deepfake detection databases and strong attribute imbalance} - The investigated Deepfake detection databases (Celeb-DF, DFD, FF++, and DFDC) lack diversity for most analysed attributes. Future works should aim to provide more unbiased, balanced, and diverse datasets to prevent the development of potentially biased Deepfake detection algorithms.
\item \textbf{\rere{Strongly correlating attribute pairs in current Deepfake detection databases}} - Future works using these databases (or our annotations) should take into account that some attributes show strong pairwise correlations to prevent misinterpretations in their results.
\item \textbf{Deepfake detection backbone networks and demographic/non-demographic attributes} - The results demonstrate that the analysed Deepfake detection backbone models are strongly biased for a variety of demographic and non-demographic attributes. The variation of the biased performances across the models and databases indicates bias \Rev{possibly originating} from several sources such as imbalanced training data, the utilised network, and their training process. 
\Rev{Low generalizability of current Deepfake detection methods can also be attributed to these omnipresent biases or imbalanced attributes. We expect bias-mitigating Deepfake detection solutions in  future work can also improve the generalizability.}
\item \textbf{Bias due to \Rev{imbalance in} attributes for pristine and Deepfake data} - For many of the investigated attributes, the biased performance similarly affects the pristine and Deepfake data. However, also the strong opposite behaviour was observed for many attributes leading the models to learn potentiality wrong \Rev{patterns}.
\item \textbf{Deepfake detection backbone models and questionable assumptions} - The results suggest that the models tend to learn questionable assumptions where the presence of a certain attribute, such as if the person is smiling or wears a hat, is a strong indicator for Deepfake. \Rev{Although this could have originated due to training data distribution}, our analysis is limited and indicates it as a potential topic in future works to enhance the reliability of these systems.
\item \textbf{Deepfake detection backbone models and societal security} - The presence of a certain attribute in a Deepfake image resulted in an increased error rate, several times higher than for a Deepfake without this attribute. Attackers \Rev{can} likely exploit these issues to increase their chances of overcoming Deepfake detection if unaddressed. On the other hand, the strong performance differences based on the presence of an attribute show a strong unfairness of these models. Future works therefore should focus on mitigating bias problems for Deepfake detection for the sake of security and society.
\end{itemize}

Based on the key observations of the three backbone networks analysed, there appears to be a significant research gap in developing Deepfake detection methods suitable for real-world applications. However, further analysis of additional methods may be necessary to make a more definitive statement.
Our analysis points to a need for more diverse and richly annotated databases for training and testing, as well as developing bias-mitigating Deepfake detection approaches.

\subsection{\Rev{Limitations Of Our Analysis}}
\Rev{While our study reveals bias issues in Deepfake detection datasets and AI-based detectors, it is essential to note the difference between correlation and causation in our analysis. Our results demonstrate strong correlations between attributes and biased performance in detection, but they do not necessarily establish causation. Bias in datasets arises from various complex factors, including data collection methodologies, historical biases, and societal contexts. While we provide valuable insights into the presence of bias, further research is needed to ascertain the causative factors responsible for these biases. This understanding is crucial for designing effective strategies to mitigate bias in Deepfake detection and to develop more equitable and reliable detectors. Further, our work does not analyse all available state-of-the-art detection approaches leaving an open question on architecture dependence and fairness factors.}

\section{Conclusion}

In this work, we provided large-scale annotations for five popular Deepfake detection datasets and used these to comprehensively analyse bias in Deepfake detection.
While existing Deepfake detection databases are only sparsely annotated, we closed this gap by making over 65.3M annotations of 47 different attributes for five Deepfake detection datasets publicly available.
Based on these datasets, we comprehensively analyse bias-causing factors in Deepfake detection purely from an attribute perspective.
The results indicated that both the datasets as well as the state-of-the-art AI-based Deepfake detectors trained on this data, demonstrate strong bias issues for many demographic and non-demographic attributes.
Depending on the use case, the biased performance can result in serious societal fairness and security problems. 
Moreover, imbalanced attributes in these datasets can further lead to generalisation problems across different attributes in current Deepfake detection algorithms.
Our findings from the study and proposed publicly-available annotations are expected to help future works to effectively evaluate and mitigate bias issues in Deepfake detection and thus, to develop reliable Deepfake detectors.

\section*{Acknowledgment}
Parts of this work was carried out during the tenure of an ERCIM ’Alain Bensoussan‘ Fellowship Programme.

\section*{Statement of Ethical Use of Datasets}
\rere{We affirm that all data utilized in this study from public databases adhere to ethical guidelines, and no special ethical clearance is applicable for publicly available information.}

\ifCLASSOPTIONcaptionsoff
  \newpage
\fi

\vspace{-2mm}

{\small
\bibliographystyle{ieee}
\bibliography{egbib}
}

\newpage
\begin{IEEEbiography}[{\includegraphics[width=1in,height=1.25in,clip,keepaspectratio]{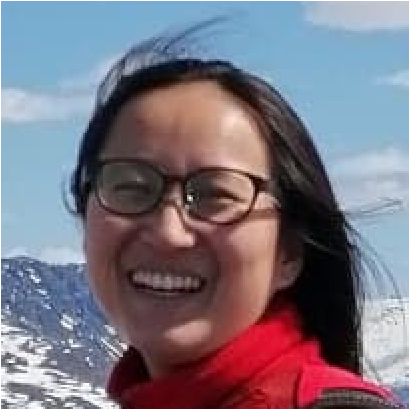}}]{Ying Xu}
received her B.Sc. in Electrical Engineering in 2015 from Shanghai University, China. She completed her M.Sc. in Applied Computer Science in 2021 from Norwegian University of Science and Technology, Norway, where she is currently pursuing the Ph.D. degree, focusing on Deepfake detection.
\end{IEEEbiography}

\vspace{11pt}

\begin{IEEEbiography}[{\includegraphics[width=1in,height=1.25in,clip,keepaspectratio]{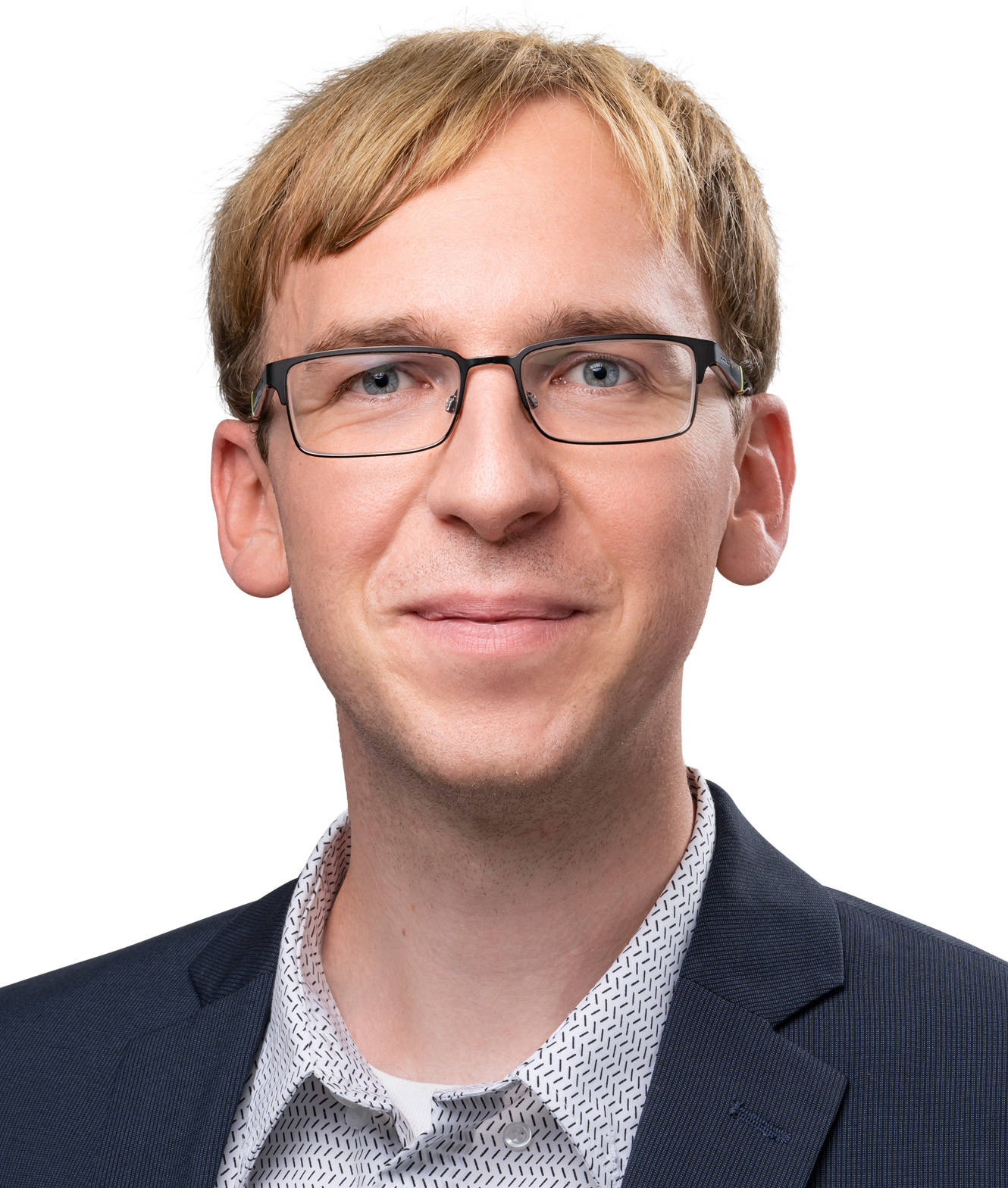}}]{Philipp Terhörst}
graduated with a Master of Science degree in physics from the Technical University of Darmstadt in 2017. Since then, he has been working in the Smart Living \& Biometric Technologies department at the Fraunhofer Institute for Computer Graphics Research (IGD) as a research scientist and as a PhD student at the Technical University of Darmstadt. He received his PhD in computer science in 2021 for his work on "Mitigating Soft-Biometric Driven Bias and Privacy Concerns in Face Recognition Systems".  His areas of specialization include topics in machine learning as well as biometric face recognition with a focus on quality assessment, privacy, and fairness. Dr. Terhörst is the author of several publications in conferences and journals such as CVPR and IEEE TIFS and regularly works as a reviewer for e.g. TPAMI, TIP, PR, BTAS, ICB. For his scientific work, he received several awards such as the 2020 EAB Biometrics Industry Award from the European Association for Biometrics for his dissertation or the IJCB 2020 Qualcomm PC Chairs Choice Best Student Paper Award. He furthermore participated in the ’Software Campus’ Program, a management program of the German Federal Ministry of Education and Research (BMBF).
\end{IEEEbiography}

\vspace{11pt}

\begin{IEEEbiography}[{\includegraphics[width=1in,height=1.25in,clip,keepaspectratio]{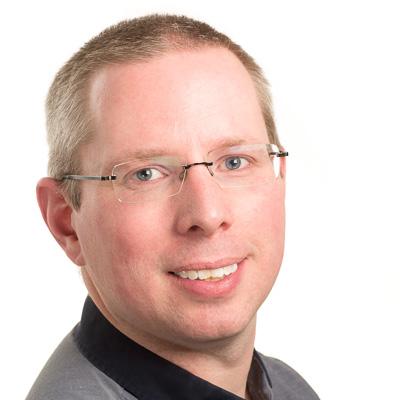}}]{Marius Pedersen}
received his BsC in Computer Engineering in 2006, and MiT in Media Technology in 2007, both from Gjøvik University College, Norway. He completed a PhD program in color imaging in 2011 from the University of Oslo, Norway, sponsored by Océ. He is professor at the Department of Computer Science at NTNU in Gjøvik, Norway. He is also the director of the Norwegian Colour and Visual Computing Laboratory (Colourlab). His work is centered on subjective and objective image quality.
\end{IEEEbiography}

\vspace{11pt}

\begin{IEEEbiography}[{\includegraphics[width=1in,height=1.25in,clip,keepaspectratio]{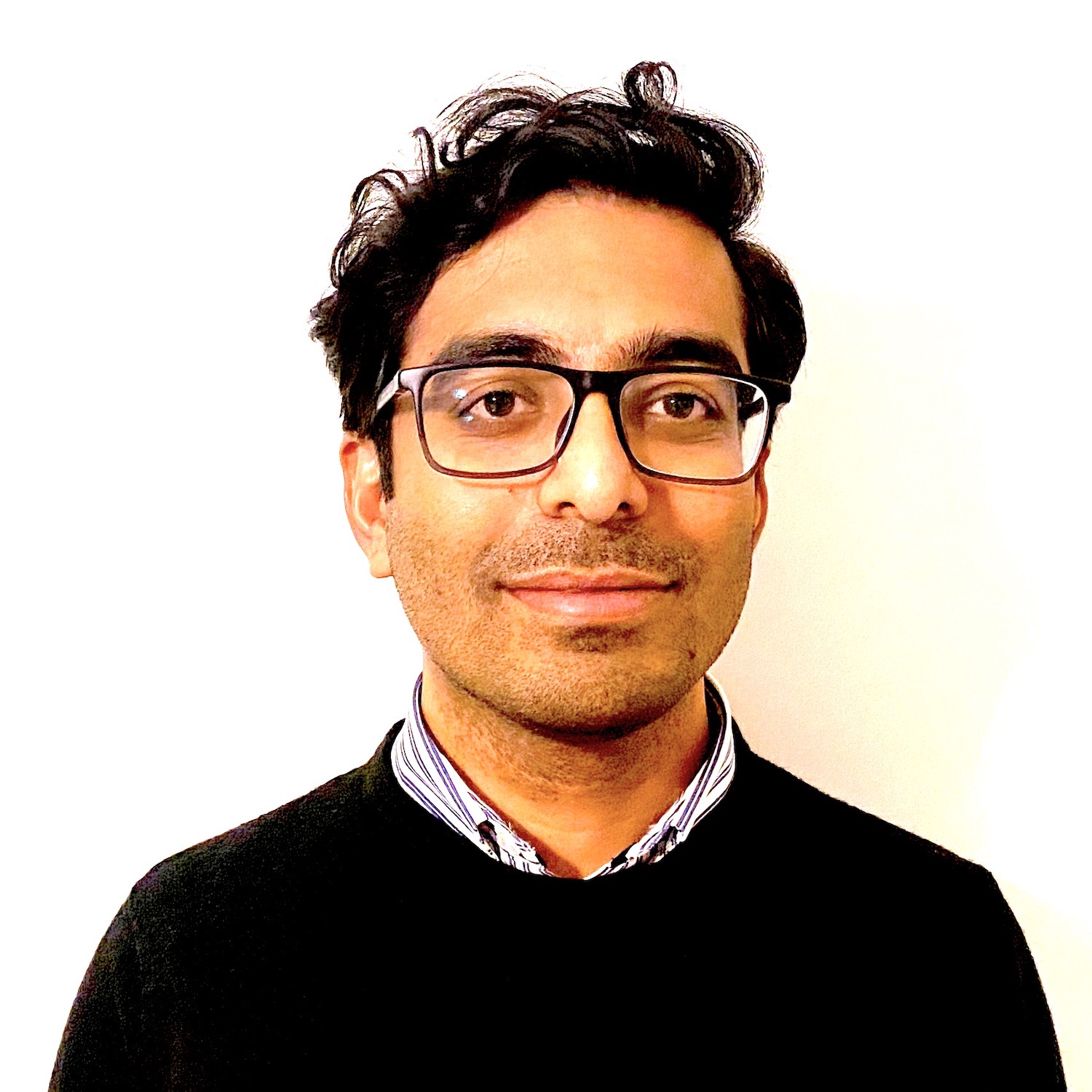}}]{Kiran Raja} (Senior Member, IEEE) received the Ph.D. degree in computer Science from the Norwegian University of Science and Technology, Norway, in 2016, where he is Faculty Member with the Department of Computer Science. His main research interests include statistical pattern recognition, image processing, and machine learning with applications to biometrics, security and privacy protection. He was/is participating in EU projects SOTAMD, iMARS, and other national projects. He is a member of the European Association of Biometrics (EAB), chairs the Academic Special Interest Group at EAB and section chair of IEEE Norway. 
\end{IEEEbiography}

\vfill


\clearpage
\newpage

\section*{Appendix}

Reporting the results of our comprehensive bias investigation resulted in a large amount of results.
The main paper focused on reporting these results in a condensed manner (e.g. in terms of relative performance measures) that allow a more easy interpretation and distillation of knowledge.
Since the detailed results of our analysis might be interesting for the research community, we will use this supplementary material to report the results in more detail.

To analyse bias originating from specific attributes, we employed a filtering process that excluded attributes with fewer than 100 positive or negative labelled images. As a result, only 31 out of the 47 available attributes in the annotated databases were included in the bias analysis. A detailed account of this process can be found in \Cref{tab:attributes_removal}, which lists all 47 attributes across 5 annotated datasets and denotes those with less than 100 labeled images using the symbol "x". Attributes that were deemed insignificant for more than two datasets were eliminated from the analysis, resulting in the selection of 31 attributes for further study. Furthermore, due to their low annotation accuracy of 0.68 as indicated in \Cref{tab:AnnotationCorrectness}, we decided to exclude the attributes of \textit{Brown Eyes} and \textit{Bags Under Eyes}.

More precisely, we report the balanced detection error as well as the error on the pristine and Deepfake data for each of the 47 attributes on five datases using three state-of-the-art Deepfake detection backbone models.
As explained in Section \ref{sec:ControlGroups}, the performance of the positive and negative groups of the data is reported as well as of the control groups.
This leads to a large amount of results that we report 15 pages of tables in this supplementary.
These are shown from Table \ref{tab:Celeb-effb0} to Table \ref{tab:DFDC-capnet}.
Each table refers to the combination of one (out of three) models with one (out of five) databases.
For attributes that were neglected for the experiments due to the low number of labels (see Section \ref{DatabaseAndConsideredAttributes}) no results are shown to avoid unreliable statements.
In some cases, the detector predicted all samples wrongly (shown as "all wrong") or correctly (shown as "all correct").
The latter case happens constantly for the DF-1.0 database.
This can be seen in Tables \ref{tab:Celeb-effb0},\ref{tab:Deeperf-xception}, and \ref{tab:Deeperf-capnet}.
Since the detectors are barely not making any errors, not enough information was provided to study bias on this databases (it was too easy).
Therefore, we neglected this database from our experiments as described in Section \ref{sec:Introduction}.

\begin{table}[t]
\centering
\setlength{\tabcolsep}{2pt}
\renewcommand{\arraystretch}{1.26}
\caption{\textbf{Metrics for identifying pertinent attributes.} The table includes all the 47 attributes in 5 annotated datasets and marks those that have less than 100 positive or negative labelled images with \textbf{x}. Attributes not deemed valuable for more than two datasets are excluded from the analysis, resulting in the selection of 31 attributes for further study. In addtion, attributes Brown Eyes and Bags Under Eyes are excluded due to low annotation accuracy.}
\label{tab:attributes_removal}
\begin{tabular}{lccccc}
\toprule
 & A-Celeb-DF & A-DFD & A-FF++ & A-DF-1.0 & A-DFDC \\ \hline
Male &  &  &  &  &  \\
Young & x &  &  &  &  \\
\rowcolor[HTML]{D9D9D9} Middle Aged & x & x &  & x & \\
\rowcolor[HTML]{D9D9D9} Senior & x & x &  & x &  \\
Asian & x & x &  &  &  \\
White &  &  &  & x &  \\
Black &  &  &  &  &  \\
\rowcolor[HTML]{D9D9D9} Rosy Cheeks & x & x & x & x & x \\
Shiny Skin & x &  &  &  &  \\
Bald &  & x &  & x &  \\
Wavy Hair &  & x &  & x & x \\
Receding Hairline &  &  &  &  &  \\
Bangs &  &  &  &  &  \\
\rowcolor[HTML]{D9D9D9} Sideburns & x &  & x & x &  \\
Black Hair &  &  &  & x &  \\
Blond Hair &  &  &  & x &  \\
\rowcolor[HTML]{D9D9D9} Brown Hair & x & x & x & x & x \\
\rowcolor[HTML]{D9D9D9} Gray Hair & x & x &  & x &  \\
No Beard &  &  &  & x &  \\
Mustache &  &  &  & x &  \\
\rowcolor[HTML]{D9D9D9} 5'o Clock Shadow & x & x & x & x &  \\
Goatee &  & x &  & x &  \\
Oval Face & x &  &  & x &  \\
Square Face &  &  &  &  &  \\
\rowcolor[HTML]{D9D9D9} Round Face & x & x & x & x &  \\
Double Chin &  &  &  & x &  \\
\rowcolor[HTML]{D9D9D9} High Cheekbones & x & x & x & x &  \\
Chubby &  &  &  &  &  \\
Obstructed Forehead &  &  &  &  &  \\
Fully Visible Forehead & x & x &  &  &  \\
\rowcolor[HTML]{F2F2F2}Brown Eyes &  &  &  & x &  \\
\rowcolor[HTML]{F2F2F2}Bags Under Eyes &  &  &  &  &  \\
\rowcolor[HTML]{D9D9D9} Bushy Eyebrows & x & x &  & x &  \\
\rowcolor[HTML]{D9D9D9} Arched Eyebrows & x & x &  & x &  \\
Mouth Closed &  &  &  &  &  \\
Smiling &  &  &  &  &  \\
Big Lips &  &  &  &  &  \\
Big Nose &  &  &  &  &  \\
Pointy Nose &  &  &  &  &  \\
Heavy Makeup &  & x &  & x & x \\
Wearing Hat &  & x &  &  &  \\
\rowcolor[HTML]{D9D9D9} Wearing Earrings & x & x & x & x & x \\
\rowcolor[HTML]{D9D9D9} Wearing Necktie & x & x &  & x &  \\
Wearing Lipstick &  & x &  & x &  \\
\rowcolor[HTML]{D9D9D9} No Eyewear & x & x &  & x &  \\
Eyeglasses &  &  &  &  &  \\
Attractive &  & x &  & x & x \\
\bottomrule
\end{tabular}
\end{table}

\begin{table*}[]
\renewcommand{\arraystretch}{0.6}
\centering
\caption{EfficientNetB0/Celeb-DF - Experiments with EfficientNetB0 on the Celeb-DF dataset.}
\label{tab:Celeb-effb0}
\begin{tabular}{lllrrrrrr}
\Xhline{2\arrayrulewidth}
Category              & Attribute                         & Class  & \multicolumn{2}{c}{Balanced Error} & \multicolumn{2}{c}{Fake Error} & \multicolumn{2}{c}{Pristine Error} \\
\cmidrule(rl){4-5} \cmidrule(rl){6-7} \cmidrule(rl){8-9}
              &                          &            & \multicolumn{1}{c}{Data}        & \multicolumn{1}{c}{Control}   & \multicolumn{1}{c}{Data}               & \multicolumn{1}{c}{Control}          & \multicolumn{1}{c}{Data}               & \multicolumn{1}{c}{Control}          \\
              \hline
Demographics  & Male                     & Positive   & 15.01\%                  & 9.34\%                      & 6.87\%                   & 2.94\%                      & 23.16\%                  & 16.76\%                     \\
              &                          & Negative   & 7.79\%                   & 10.09\%                     & 3.05\%                   & 3.59\%                      & 12.53\%                  & 16.59\%                     \\
              &                          & Rel. Perf. & -92.70\%                 & 7.41\%                      & -125.36\%                & 18.07\%                     & -84.76\%                 & -1.01\%                     \\
              & Young                    & Positive   & -                        & -                           & -                        & -                           & -                        & -                           \\
              &                          & Negative   & -                        & -                           & -                        & -                           & -                        & -                           \\
              &                          & Rel. Perf. & -                        & -                           & -                        & -                           & -                        & -                           \\
\end{tabular}
\end{table*}              

\begin{table*}[]
\renewcommand{\arraystretch}{0.6}
\centering
\caption{Xception/Celeb-DF - Experiments with Xception on the Celeb-DF dataset.}
\label{tab:Celeb-xception}
\begin{tabular}{lllrrrrrr}
\Xhline{2\arrayrulewidth}
Category              & Attribute                         & Class  & \multicolumn{2}{c}{Balanced Error} & \multicolumn{2}{c}{Fake Error} & \multicolumn{2}{c}{Pristine Error} \\
\cmidrule(rl){4-5} \cmidrule(rl){6-7} \cmidrule(rl){8-9}
              &                          &            & \multicolumn{1}{c}{Data}        & \multicolumn{1}{c}{Control}   & \multicolumn{1}{c}{Data}               & \multicolumn{1}{c}{Control}          & \multicolumn{1}{c}{Data}               & \multicolumn{1}{c}{Control}          \\
              \hline
Demographics  & Male                     & Positive   & 12.54\%                  & 12.13\%                     & 13.09\%                  & 15.65\%                     & 11.98\%                  & 7.81\%                      \\
              &                          & Negative   & 9.67\%                   & 11.76\%                     & 15.98\%                  & 15.57\%                     & 3.36\%                   & 7.52\%                      \\
              &                          & Rel. Perf. & -29.63\%                 & -3.11\%                     & 18.09\%                  & -0.54\%                     & -256.25\%                & -3.85\%                     \\
              & Young                    & Positive   & -                        & -                           & -                        & -                           & -                        & -                           \\
              &                          & Negative   & -                        & -                           & -                        & -                           & -                        & -                           \\
              &                          & Rel. Perf. & -                        & -                           & -                        & -                           & -                        & -                           \\
\end{tabular}
\end{table*}              

\begin{table*}[]
\renewcommand{\arraystretch}{0.6}
\centering
\caption{Capsule-Forensics-v2/Celeb-DF - Experiments with Capsule-Forensics-v2 on the Celeb-DF dataset.}
\label{tab:Celeb-meso}
\begin{tabular}{lllrrrrrr}
\Xhline{2\arrayrulewidth}
Category              & Attribute                         & Class  & \multicolumn{2}{c}{Balanced Error} & \multicolumn{2}{c}{Fake Error} & \multicolumn{2}{c}{Pristine Error} \\
\cmidrule(rl){4-5} \cmidrule(rl){6-7} \cmidrule(rl){8-9}
              &                          &            & \multicolumn{1}{c}{Data}        & \multicolumn{1}{c}{Control}   & \multicolumn{1}{c}{Data}               & \multicolumn{1}{c}{Control}          & \multicolumn{1}{c}{Data}               & \multicolumn{1}{c}{Control}          \\
              \hline
Demographics  & Male                     & Positive   & 13.73\%                  & 12.25\%                     & 3.65\%                   & 2.18\%                      & 23.81\%                  & 22.83\%                     \\
              &                          & Negative   & 11.96\%                  & 12.43\%                     & 1.92\%                   & 2.07\%                      & 22.00\%                  & 22.63\%                     \\
              &                          & Rel. Perf. & -14.79\%                 & 1.38\%                      & -89.98\%                 & -5.39\%                     & -8.22\%                  & -0.91\%                     \\
              & Young                    & Positive   & -                        & -                           & -                        & -                           & -                        & -                           \\
              &                          & Negative   & -                        & -                           & -                        & -                           & -                        & -                           \\
              &                          & Rel. Perf. & -                        & -                           & -                        & -                           & -                        & -                           \\
\end{tabular}
\end{table*}

\begin{table*}[]
\renewcommand{\arraystretch}{0.6}
\centering
\caption{EfficientNetB0/DFD - Experiments with EfficientNetB0 on the DFD dataset.}
\label{tab:DFD-effb0}
\begin{tabular}{lllrrrrrr}
\Xhline{2\arrayrulewidth}
Category              & Attribute                         & Class  & \multicolumn{2}{c}{Balanced Error} & \multicolumn{2}{c}{Fake Error} & \multicolumn{2}{c}{Pristine Error} \\
\cmidrule(rl){4-5} \cmidrule(rl){6-7} \cmidrule(rl){8-9}
              &                          &            & \multicolumn{1}{c}{Data}        & \multicolumn{1}{c}{Control}   & \multicolumn{1}{c}{Data}               & \multicolumn{1}{c}{Control}          & \multicolumn{1}{c}{Data}               & \multicolumn{1}{c}{Control}          \\
              \hline
Demographics  & Male                     & Positive   & 0.71\%                   & 2.85\%                      & 0.28\%                   & 1.67\%                      & 6.77\%                   & 8.58\%                      \\
              &                          & Negative   & 4.13\%                   & 2.78\%                      & 2.88\%                   & 1.67\%                      & 10.80\%                  & 9.04\%                      \\
              &                          & Rel. Perf. & 82.79\%                  & -2.34\%                     & 90.43\%                  & -0.29\%                     & 37.30\%                  & 5.11\%                      \\
              & Young                    & Positive   & 3.07\%                   & 2.97\%                      & 1.06\%                   & 1.79\%                      & 12.26\%                  & 8.91\%                      \\
              &                          & Negative   & 3.24\%                   & 3.67\%                      & 3.98\%                   & 1.86\%                      & 0.00\%                   & 4.65\%                      \\
              &                          & Rel. Perf. & 5.16\%                   & 19.18\%                     & 73.44\%                  & 3.43\%                      & All wrong                & -91.46\%                    \\
\end{tabular}
\end{table*}              

\begin{table*}[]
\renewcommand{\arraystretch}{0.6}
\centering
\caption{Xception/DFD - Experiments with Xception on the DFD dataset.}
\label{tab:DFD-xception}
\begin{tabular}{lllrrrrrr}
\Xhline{2\arrayrulewidth}
Category              & Attribute                         & Class  & \multicolumn{2}{c}{Balanced Error} & \multicolumn{2}{c}{Fake Error} & \multicolumn{2}{c}{Pristine Error} \\
\cmidrule(rl){4-5} \cmidrule(rl){6-7} \cmidrule(rl){8-9}
              &                          &            & \multicolumn{1}{c}{Data}        & \multicolumn{1}{c}{Control}   & \multicolumn{1}{c}{Data}               & \multicolumn{1}{c}{Control}          & \multicolumn{1}{c}{Data}               & \multicolumn{1}{c}{Control}          \\
              \hline
Demographics  & Male                     & Positive   & 6.41\%                   & 5.75\%                      & 0.41\%                   & 3.23\%                      & 12.42\%                  & 6.32\%                      \\
              &                          & Negative   & 6.32\%                   & 5.49\%                      & 5.60\%                   & 3.21\%                      & 7.04\%                   & 7.28\%                      \\
              &                          & Rel. Perf. & -1.43\%                  & -4.73\%                     & 92.76\%                  & -0.50\%                     & -76.35\%                 & 13.18\%                     \\
              & Young                    & Positive   & 4.50\%                   & 5.92\%                      & 3.46\%                   & 3.43\%                      & 5.55\%                   & 7.01\%                      \\
              &                          & Negative   & 14.25\%                  & 7.52\%                      & 2.92\%                   & 3.18\%                      & 25.58\%                  & 4.65\%                      \\
              &                          & Rel. Perf. & 68.40\%                  & 21.25\%                     & -18.52\%                 & -7.64\%                     & 78.31\%                  & -50.66\%                    \\
\end{tabular}
\end{table*}              

\begin{table*}[]
\renewcommand{\arraystretch}{0.6}
\centering
\caption{Capsule-Forensics-v2/DFD - Experiments with Capsule-Forensics-v2 on the DFD dataset.}
\label{tab:DFD-capnet}
\begin{tabular}{lllrrrrrr}
\Xhline{2\arrayrulewidth}
Category              & Attribute                         & Class  & \multicolumn{2}{c}{Balanced Error} & \multicolumn{2}{c}{Fake Error} & \multicolumn{2}{c}{Pristine Error} \\
\cmidrule(rl){4-5} \cmidrule(rl){6-7} \cmidrule(rl){8-9}
              &                          &            & \multicolumn{1}{c}{Data}        & \multicolumn{1}{c}{Control}   & \multicolumn{1}{c}{Data}               & \multicolumn{1}{c}{Control}          & \multicolumn{1}{c}{Data}               & \multicolumn{1}{c}{Control}          \\
              \hline
Demographics  & Male                     & Positive   & 7.89\%                   & 11.90\%                     & 0.19\%                   & 3.49\%                      & 15.58\%                  & 20.99\%                     \\
              &                          & Negative   & 14.85\%                  & 12.10\%                     & 6.17\%                   & 3.44\%                      & 23.52\%                  & 20.96\%                     \\
              &                          & Rel. Perf. & 46.88\%                  & 1.59\%                      & 96.84\%                  & -1.47\%                     & 33.78\%                  & -0.16\%                     \\
              & Young                    & Positive   & 16.37\%                  & 11.20\%                     & 3.11\%                   & 3.68\%                      & 29.64\%                  & 20.58\%                     \\
              &                          & Negative   & 15.70\%                  & 10.97\%                     & 0.00\%                   & 3.18\%                      & 31.40\%                  & 17.44\%                     \\
              &                          & Rel. Perf. & -4.29\%                  & -2.17\%                     & All correct              & -15.69\%                    & 5.61\%                   & -18.01\%                    \\
\end{tabular}
\end{table*}

\begin{table*}[]
\renewcommand{\arraystretch}{0.6}
\centering
\caption{EfficientNetB0/FF++ - Experiments with EfficientNetB0 on the FF++ dataset.}
\label{tab:FF-effb0}
\begin{tabular}{lllrrrrrr}
\Xhline{2\arrayrulewidth}
Category              & Attribute                         & Class  & \multicolumn{2}{c}{Balanced Error} & \multicolumn{2}{c}{Fake Error} & \multicolumn{2}{c}{Pristine Error} \\
\cmidrule(rl){4-5} \cmidrule(rl){6-7} \cmidrule(rl){8-9}
              &                          &            & \multicolumn{1}{c}{Data}        & \multicolumn{1}{c}{Control}   & \multicolumn{1}{c}{Data}               & \multicolumn{1}{c}{Control}          & \multicolumn{1}{c}{Data}               & \multicolumn{1}{c}{Control}          \\
              \hline
Demographics  & Male                     & Positive   & 1.63\%                   & 1.45\%                      & 1.27\%                   & 0.78\%                      & 3.37\%                   & 4.26\%                      \\
              &                          & Negative   & 1.42\%                   & 1.44\%                      & 0.65\%                   & 0.84\%                      & 5.28\%                   & 4.56\%                      \\
              &                          & Rel. Perf. & -14.66\%                 & -0.78\%                     & -96.99\%                 & 7.17\%                      & 36.12\%                  & 6.55\%                      \\
              & Young                    & Positive   & 1.12\%                   & 1.47\%                      & 0.43\%                   & 0.74\%                      & 4.55\%                   & 4.24\%                      \\
              &                          & Negative   & 0.41\%                   & 1.55\%                      & 0.00\%                   & 0.61\%                      & 2.44\%                   & 4.89\%                      \\
              &                          & Rel. Perf. & -172.49\%                & 4.66\%                      & All right                & -21.58\%                    & -86.06\%                 & 13.31\%                     \\
\end{tabular}
\end{table*}              

\begin{table*}[]
\renewcommand{\arraystretch}{0.6}
\centering
\caption{Xception/FF++ - Experiments with Xception on the FF++ dataset.}
\label{tab:FF-xception}
\begin{tabular}{lllrrrrrr}
\Xhline{2\arrayrulewidth}
Category              & Attribute                         & Class  & \multicolumn{2}{c}{Balanced Error} & \multicolumn{2}{c}{Fake Error} & \multicolumn{2}{c}{Pristine Error} \\
\cmidrule(rl){4-5} \cmidrule(rl){6-7} \cmidrule(rl){8-9}
              &                          &            & \multicolumn{1}{c}{Data}        & \multicolumn{1}{c}{Control}   & \multicolumn{1}{c}{Data}               & \multicolumn{1}{c}{Control}          & \multicolumn{1}{c}{Data}               & \multicolumn{1}{c}{Control}          \\
              \hline
Demographics  & Male                     & Positive   & 1.54\%                   & 1.24\%                      & 1.52\%                   & 1.06\%                      & 1.63\%                   & 1.74\%                      \\
              &                          & Negative   & 1.12\%                   & 1.23\%                      & 0.95\%                   & 1.13\%                      & 2.00\%                   & 1.91\%                      \\
              &                          & Rel. Perf. & -37.21\%                 & -1.45\%                     & -60.77\%                 & 6.58\%                      & 18.44\%                  & 8.59\%                      \\
              & Young                    & Positive   & 0.67\%                   & 1.21\%                      & 0.63\%                   & 1.03\%                      & 0.88\%                   & 2.01\%                      \\
              &                          & Negative   & 0.11\%                   & 1.33\%                      & 0.08\%                   & 0.91\%                      & 0.24\%                   & 2.12\%                      \\
              &                          & Rel. Perf. & -509.93\%                & 9.19\%                      & -660.29\%                & -14.04\%                    & -259.44\%                & 4.86\%                      \\
\end{tabular}
\end{table*}              

\begin{table*}[]
\renewcommand{\arraystretch}{0.6}
\centering
\caption{Capsule-Forensics-v2/FF++ - Experiments with Capsule-Forensics-v2 on the FF++ dataset.}
\label{tab:FF-capnet}
\begin{tabular}{lllrrrrrr}
\Xhline{2\arrayrulewidth}
Category              & Attribute                         & Class  & \multicolumn{2}{c}{Balanced Error} & \multicolumn{2}{c}{Fake Error} & \multicolumn{2}{c}{Pristine Error} \\
\cmidrule(rl){4-5} \cmidrule(rl){6-7} \cmidrule(rl){8-9}
              &                          &            & \multicolumn{1}{c}{Data}        & \multicolumn{1}{c}{Control}   & \multicolumn{1}{c}{Data}               & \multicolumn{1}{c}{Control}          & \multicolumn{1}{c}{Data}               & \multicolumn{1}{c}{Control}          \\
              \hline
Demographics  & Male                     & Positive   & 4.39\%                   & 4.21\%                      & 2.07\%                   & 1.67\%                      & 6.71\%                   & 6.22\%                      \\
              &                          & Negative   & 3.70\%                   & 4.33\%                      & 1.19\%                   & 1.69\%                      & 6.21\%                   & 6.25\%                      \\
              &                          & Rel. Perf. & -18.57\%                 & 2.85\%                      & -74.32\%                 & 1.05\%                      & -7.91\%                  & 0.35\%                      \\
              & Young                    & Positive   & 3.11\%                   & 4.20\%                      & 0.65\%                   & 1.65\%                      & 5.58\%                   & 6.66\%                      \\
              &                          & Negative   & 2.57\%                   & 3.61\%                      & 1.14\%                   & 1.54\%                      & 3.99\%                   & 6.76\%                      \\
              &                          & Rel. Perf. & -21.32\%                 & -16.38\%                    & 43.16\%                  & -7.18\%                     & -39.80\%                 & 1.42\%                      \\
\end{tabular}
\end{table*}

\begin{table*}[]
\renewcommand{\arraystretch}{0.6}
\centering
\caption{EfficientNetB0/DF-1.0 - Experiments with EfficientNetB0 on the DF-1.0 dataset.}
\label{tab:Deeperf-effb0}
\begin{tabular}{lllrrrrrr}
\Xhline{2\arrayrulewidth}
Category              & Attribute                         & Class  & \multicolumn{2}{c}{Balanced Error} & \multicolumn{2}{c}{Fake Error} & \multicolumn{2}{c}{Pristine Error} \\
\cmidrule(rl){4-5} \cmidrule(rl){6-7} \cmidrule(rl){8-9}
              &                          &            & \multicolumn{1}{c}{Data}        & \multicolumn{1}{c}{Control}   & \multicolumn{1}{c}{Data}               & \multicolumn{1}{c}{Control}          & \multicolumn{1}{c}{Data}               & \multicolumn{1}{c}{Control}          \\
              \hline
Demographics  & Male                     & Positive   & 0.07\%                   & 0.06\%                      & 0.03\%                   & 0.00\%                      & 0.11\%                   & 0.10\%                      \\
              &                          & Negative   & 0.11\%                   & 0.06\%                      & 0.00\%                   & 0.00\%                      & 0.22\%                   & 0.11\%                      \\
              &                          & Rel. Perf. & 39.32\%                  & -12.10\%                    & -821.51\%                & -31.64\%                    & 51.23\%                  & 7.24\%                      \\
              & Young                    & Positive   & 0.10\%                   & 0.05\%                      & 0.00\%                   & 0.01\%                      & 0.20\%                   & 0.11\%                      \\
              &                          & Negative   & 0.00\%                   & 0.06\%                      & 0.00\%                   & 0.00\%                      & 0.00\%                   & 0.12\%                      \\
              &                          & Rel. Perf. & All correct              & 22.42\%                     & All correct              & All correct                 & All correct              & 9.04\%                      \\
\end{tabular}
\end{table*}              

\begin{table*}[]
\renewcommand{\arraystretch}{0.6}
\centering
\caption{Xception/DF-1.0 - Experiments with Xception on the DF-1.0 dataset.}
\label{tab:Deeperf-xception}
\begin{tabular}{lllrrrrrr}
\Xhline{2\arrayrulewidth}
Category              & Attribute                         & Class  & \multicolumn{2}{c}{Balanced Error} & \multicolumn{2}{c}{Fake Error} & \multicolumn{2}{c}{Pristine Error} \\
\cmidrule(rl){4-5} \cmidrule(rl){6-7} \cmidrule(rl){8-9}
              &                          &            & \multicolumn{1}{c}{Data}        & \multicolumn{1}{c}{Control}   & \multicolumn{1}{c}{Data}               & \multicolumn{1}{c}{Control}          & \multicolumn{1}{c}{Data}               & \multicolumn{1}{c}{Control}          \\
              \hline
Demographics  & Male                     & Positive   & 0.01\%                   & 0.01\%                      & 0.02\%                   & 0.02\%                      & 0.00\%                   & 0.00\%                      \\
              &                          & Negative   & 0.00\%                   & 0.00\%                      & 0.01\%                   & 0.01\%                      & 0.00\%                   & 0.00\%                      \\
              &                          & Rel. Perf. & -163.29\%                & -118.88\%                   & -163.29\%                & -31.64\%                    & All correct              & All correct                 \\
              & Young                    & Positive   & 0.00\%                   & 0.00\%                      & 0.01\%                   & 0.02\%                      & 0.00\%                   & 0.00\%                      \\
              &                          & Negative   & 0.00\%                   & 0.00\%                      & 0.00\%                   & 0.00\%                      & 0.00\%                   & 0.00\%                      \\
              &                          & Rel. Perf. & All correct              & All correct                 & All correct              & All correct                 & All correct              & All correct                 \\
\end{tabular}
\end{table*}              

\begin{table*}[]
\renewcommand{\arraystretch}{0.6}
\centering
\caption{Capsule-Forensics-v2/DF-1.0 - Experiments with Capsule-Forensics-v2 on the DF-1.0 dataset.}
\label{tab:Deeperf-capnet}
\begin{tabular}{lllrrrrrr}
\Xhline{2\arrayrulewidth}
Category              & Attribute                         & Class  & \multicolumn{2}{c}{Balanced Error} & \multicolumn{2}{c}{Fake Error} & \multicolumn{2}{c}{Pristine Error} \\
\cmidrule(rl){4-5} \cmidrule(rl){6-7} \cmidrule(rl){8-9}
              &                          &            & \multicolumn{1}{c}{Data}        & \multicolumn{1}{c}{Control}   & \multicolumn{1}{c}{Data}               & \multicolumn{1}{c}{Control}          & \multicolumn{1}{c}{Data}               & \multicolumn{1}{c}{Control}          \\
              \hline
Demographics  & Male                     & Positive   & 0                        & 0                           & 0                        & 0                           & 0                        & 0                           \\
              &                          & Negative   & 0                        & 0                           & 0                        & 0                           & 0                        & 0                           \\
              &                          & Rel. Perf. & All correct              & -35.40\%                    & All correct              & All correct                 & All correct              & All correct                 \\
              & Young                    & Positive   & 0                        & 0                           & 0                        & 0                           & 0                        & 0                           \\
              &                          & Negative   & 0                        & 0                           & 0                        & 0                           & 0                        & 0                           \\
              &                          & Rel. Perf. & All correct              & All correct                 & All correct              & All correct                 & All correct              & All correct                 \\
\end{tabular}
\end{table*}

\begin{table*}[]
\renewcommand{\arraystretch}{0.6}
\centering
\caption{EfficientNetB0/DFDC - Experiments with EfficientNetB0 on the DFDC dataset.}
\label{tab:DFDC-effb0}
\begin{tabular}{lllrrrrrr}
\Xhline{2\arrayrulewidth}
Category              & Attribute                         & Class  & \multicolumn{2}{c}{Balanced Error} & \multicolumn{2}{c}{Fake Error} & \multicolumn{2}{c}{Pristine Error} \\
\cmidrule(rl){4-5} \cmidrule(rl){6-7} \cmidrule(rl){8-9}
              &                          &            & \multicolumn{1}{c}{Data}        & \multicolumn{1}{c}{Control}   & \multicolumn{1}{c}{Data}               & \multicolumn{1}{c}{Control}          & \multicolumn{1}{c}{Data}               & \multicolumn{1}{c}{Control}          \\
              \hline
Demographics  & Male                     & Positive   & 24.12\%                  & 23.66\%                     & 13.00\%                  & 11.89\%                     & 35.25\%                  & 35.52\%                     \\
              &                          & Negative   & 21.88\%                  & 23.66\%                     & 10.18\%                  & 11.83\%                     & 33.58\%                  & 35.68\%                     \\
              &                          & Rel. Perf. & -10.26\%                 & -0.01\%                     & -27.77\%                 & -0.52\%                     & -4.95\%                  & 0.46\%                      \\
              & Young                    & Positive   & 17.66\%                  & 23.57\%                     & 9.25\%                   & 10.93\%                     & 26.06\%                  & 35.93\%                     \\
              &                          & Negative   & 22.33\%                  & 23.50\%                     & 13.55\%                  & 11.54\%                     & 31.11\%                  & 35.76\%                     \\
              &                          & Rel. Perf. & 20.92\%                  & -0.29\%                     & 31.72\%                  & 5.27\%                      & 16.22\%                  & -0.48\%                     \\
\end{tabular}
\end{table*}              

\begin{table*}[]
\renewcommand{\arraystretch}{0.6}
\centering
\caption{Xception/DFDC - Experiments with Xception on the DFDC dataset.}
\label{tab:DFDC-xception}
\begin{tabular}{lllrrrrrr}
\Xhline{2\arrayrulewidth}
Category              & Attribute                         & Class  & \multicolumn{2}{c}{Balanced Error} & \multicolumn{2}{c}{Fake Error} & \multicolumn{2}{c}{Pristine Error} \\
\cmidrule(rl){4-5} \cmidrule(rl){6-7} \cmidrule(rl){8-9}
              &                          &            & \multicolumn{1}{c}{Data}        & \multicolumn{1}{c}{Control}   & \multicolumn{1}{c}{Data}               & \multicolumn{1}{c}{Control}          & \multicolumn{1}{c}{Data}               & \multicolumn{1}{c}{Control}          \\
              \hline
Demographics  & Male                     & Positive   & 28.83\%                  & 27.98\%                     & 21.35\%                  & 17.42\%                     & 36.31\%                  & 38.65\%                     \\
              &                          & Negative   & 25.73\%                  & 28.00\%                     & 12.36\%                  & 17.31\%                     & 39.10\%                  & 38.72\%                     \\
              &                          & Rel. Perf. & -12.05\%                 & 0.09\%                      & -72.74\%                 & -0.62\%                     & 7.13\%                   & 0.20\%                      \\
              & Young                    & Positive   & 22.95\%                  & 27.40\%                     & 9.56\%                   & 17.16\%                     & 36.34\%                  & 38.91\%                     \\
              &                          & Negative   & 27.93\%                  & 27.38\%                     & 20.47\%                  & 17.28\%                     & 35.40\%                  & 38.37\%                     \\
              &                          & Rel. Perf. & 17.84\%                  & -0.07\%                     & 53.31\%                  & 0.74\%                      & -2.67\%                  & -1.41\%                     \\
\end{tabular}
\end{table*}              

\begin{table*}[]
\renewcommand{\arraystretch}{0.6}
\centering
\caption{Capsule-Forensics-v2/DFDC - Experiments with Capsule-Forensics-v2 on the DFDC dataset.}
\label{tab:DFDC-capnet}
\begin{tabular}{lllrrrrrr}
\Xhline{2\arrayrulewidth}
Category              & Attribute                         & Class  & \multicolumn{2}{c}{Balanced Error} & \multicolumn{2}{c}{Fake Error} & \multicolumn{2}{c}{Pristine Error} \\
\cmidrule(rl){4-5} \cmidrule(rl){6-7} \cmidrule(rl){8-9}
              &                          &            & \multicolumn{1}{c}{Data}        & \multicolumn{1}{c}{Control}   & \multicolumn{1}{c}{Data}               & \multicolumn{1}{c}{Control}          & \multicolumn{1}{c}{Data}               & \multicolumn{1}{c}{Control}          \\
              \hline
Demographics  & Male                     & Positive   & 27.64\%                  & 26.80\%                     & 12.24\%                  & 10.91\%                     & 43.05\%                  & 43.09\%                     \\
              &                          & Negative   & 24.15\%                  & 26.76\%                     & 9.06\%                   & 10.86\%                     & 39.23\%                  & 43.27\%                     \\
              &                          & Rel. Perf. & -14.48\%                 & -0.18\%                     & -35.12\%                 & -0.49\%                     & -9.71\%                  & 0.41\%                      \\
              & Young                    & Positive   & 19.74\%                  & 27.38\%                     & 8.67\%                   & 10.32\%                     & 30.82\%                  & 43.61\%                     \\
              &                          & Negative   & 27.38\%                  & 27.08\%                     & 15.15\%                  & 10.98\%                     & 39.61\%                  & 43.82\%                     \\
              &                          & Rel. Perf. & 27.88\%                  & -1.10\%                     & 42.76\%                  & 5.95\%                      & 22.19\%                  & 0.46\%                      \\
\end{tabular}
\end{table*}

\end{document}